\documentclass{article}

\PassOptionsToPackage{authoryear,round}{natbib}
\usepackage[preprint]{neurips_2026}
\makeatletter
\renewcommand{\@notice}{\enlargethispage{2\baselineskip}}
\makeatother

\usepackage[utf8]{inputenc}
\usepackage[T1]{fontenc}

\usepackage{microtype}

\usepackage{graphicx}
\usepackage{subcaption}
\usepackage{booktabs} 
\usepackage{colortbl}
\usepackage{hyperref}
\usepackage{url}
\usepackage{xcolor}
\usepackage{caption}
\usepackage{float}
\usepackage{adjustbox}

\usepackage{amsmath}
\usepackage{amssymb}
\usepackage{amsfonts}
\usepackage{mathtools}
\usepackage{amsthm}

\usepackage[capitalize,noabbrev]{cleveref}

\theoremstyle{plain}

\theoremstyle{definition}

\theoremstyle{remark}

\title{From Representational Complementarity to Dual Systems: Synergizing VLM and Vision-Only Backbones for End-to-End Driving}

\author{%
  \textbf{Sining Ang}\textsuperscript{1,2} \quad
  \textbf{Yuguang Yang}\textsuperscript{1,3} \quad
  \textbf{Chenxu Dang}\textsuperscript{1,4} \quad
  \textbf{Canyu Chen}\textsuperscript{1,5} \quad
  \textbf{Cheng Chi}\textsuperscript{6} \\
  \textbf{Haiyan Liu}\textsuperscript{7} \quad
  \textbf{Xuanyao Mao}\textsuperscript{7} \quad
  \textbf{Jason Bao}\textsuperscript{7} \quad
  \textbf{Xuliang}\textsuperscript{7} \quad
  \textbf{Bingchuan Sun}\textsuperscript{7} \quad
  \textbf{Yan Wang}\textsuperscript{1} \\[2pt]
  \footnotesize
  \textsuperscript{1}Institute for AI Industry Research (AIR), Tsinghua University \\
  \textsuperscript{2}University of Science and Technology of China \quad
  \textsuperscript{3}Beihang University \\
  \textsuperscript{4}Huazhong University of Science and Technology \\
  \textsuperscript{5}National Superior College for Engineers, Beihang University \\
  \textsuperscript{6}Beijing Academy of Artificial Intelligence \quad
  \textsuperscript{7}Lenovo Group Limited
}
\hypersetup{hidelinks,pdfauthor={Sining Ang; Yuguang Yang; Chenxu Dang; Canyu Chen; Cheng Chi; Haiyan Liu; Xuanyao Mao; Jason Bao; Xuliang; Bingchuan Sun; Yan Wang}}

\begin{document}

\maketitle

\begin{abstract}
Vision-Language-Action (VLA) driving augments end-to-end (E2E) planning with language-enabled visual backbones, but how VLM representations differ from vision-only encoders after policy learning, and whether such differences matter for planning, remains unclear. Under a unified VLM-hidden + diffusion-policy paradigm, we compare multiple VLM families/scales (InternVL3 and Qwen3VL) with standard vision-only encoders (ResNet, ViT, and EVA-CLIP) while keeping the downstream planner fixed. We study representation, behavior, and system design. CKA/CCA and Shared--Unique SAE show that policy learning enlarges a common decision subspace, but both branches retain non-transferable residual factors. We further replicate this shared-plus-unique representation pattern on nuPlan using the AsyncDrive planning stack. Latent-intervention policies and scenario-level analysis further show that these residuals are behaviorally meaningful: vision-only encoders are stronger in simple geometry-dominant scenes, whereas VLMs are more effective in semantically complex and interaction-heavy long-tail cases. The two branches also exhibit distinct progress--braking and path-choice tendencies, and an oracle best-of-two VLM+ViT selector reaches 93.58 PDMS on NAVSIM. We convert this complementarity into two lightweight systems: \textbf{HybridDriveVLA}, which selects from a compact cross-model candidate set using a learned trajectory scorer and improves PDMS from 90.80 to 92.10, and \textbf{DualDriveVLA}, a fast--slow variant that invokes the VLM in only 15\% of scenarios, achieving 91.00 PDMS with about $1.9\times$ lower latency than the VLM baseline. Code will be available at \url{https://github.com/WilliamXuanYu/HybridDriveVLA}.
\end{abstract}

\section{Introduction}

End-to-end driving planning maps visual observations directly to future trajectories, reducing reliance on hand-crafted modular pipelines. 
In this setting, the visual encoder affects not only planning quality but also latency, memory, and deployment cost. 
Vision--language models (VLMs) are often believed to provide stronger semantics and long-tail reasoning than standard vision-only encoders, yet it remains unclear how the two families differ internally after policy learning, whether the differences are behaviorally meaningful, and how they can be exploited in planning.

We study this question under a unified \emph{VLM-hidden + diffusion-policy} paradigm. 
Our main experiments instantiate two separate policies with the same downstream diffusion planner and action head: one branch uses a driving-domain-adapted VLM encoder, while the other replaces it with a vision-only encoder. 
This setup enables controlled comparisons between practical VLA-style and vision-only representation pipelines while keeping the downstream planner, trajectory format, data split, and training schedule fixed. 
We evaluate multiple VLM families/scales and vision-only encoders on NAVSIM, and include additional cross-setting evidence to test whether the observed patterns persist beyond a single backbone or planner.

We organize the analysis around three questions.

\begin{figure*}[t]
\centering
\vspace{-0.5em}

\begin{minipage}[t]{0.45\textwidth}
    \vspace{0pt}
    \centering
    \includegraphics[width=\linewidth]{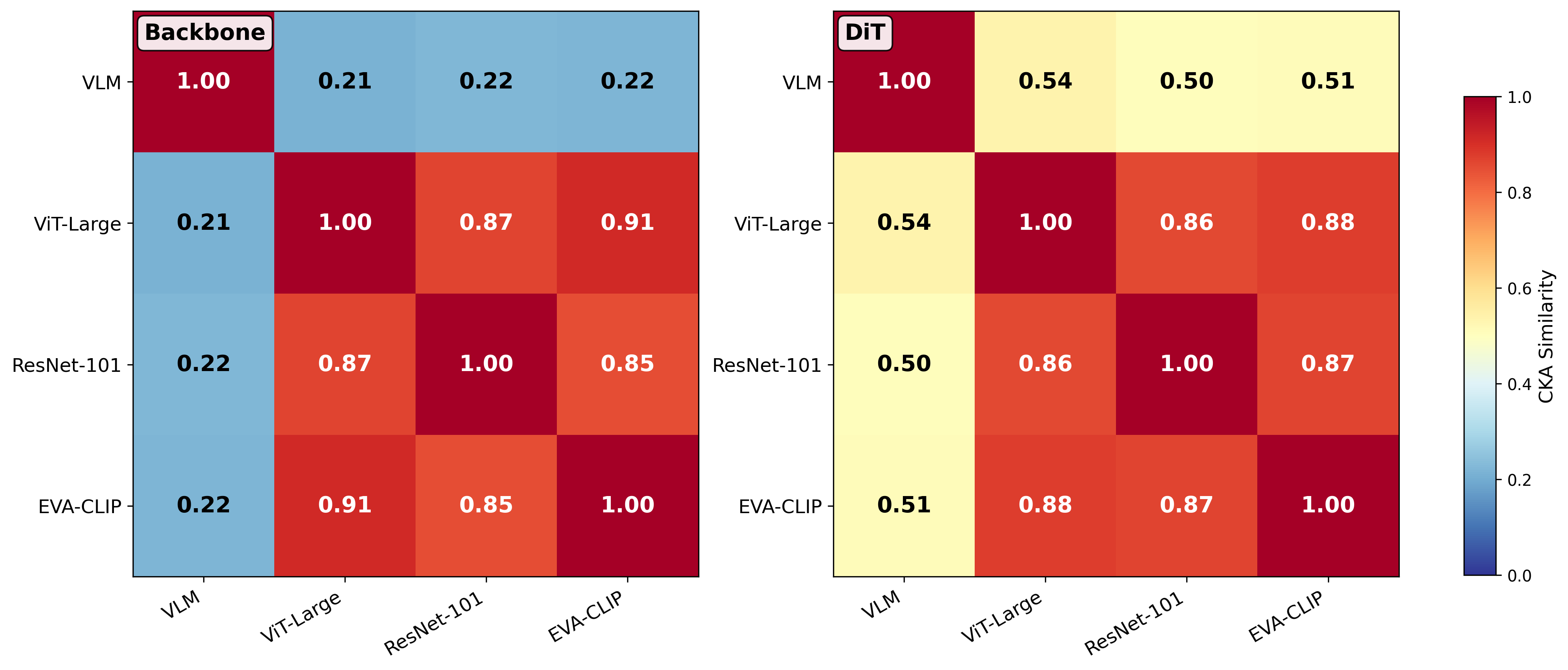}
    \captionof{figure}{
    \textbf{CKA before and after policy learning.}
    VLM--vision similarity increases at the decision level without complete collapse.
    }
    \label{fig:cka_backbone_vs_dit}
\end{minipage}
\hfill
\begin{minipage}[t]{0.52\textwidth}
    \vspace{0pt}
    \centering
    \captionof{table}{
    \textbf{RQ1 generalization.}
    The shared-plus-unique trend persists across settings.
    }
    \label{tab:rq1_generalization}
    \fontsize{8pt}{9.5pt}\selectfont
    \setlength{\tabcolsep}{3pt}
    \renewcommand{\arraystretch}{0.9}
    \begin{tabular}{@{}lccc@{}}
    \toprule
    Setting & CKA & CCA@10 & $R^2_{\mathrm{cross}}$ \\
    \midrule
    InternVL3-8B (BB) & 0.26/0.23/0.24 & 0.81 & 0.49/0.62 \\
    InternVL3-8B (Dec.) & 0.49/0.48/0.48 & 0.97 & 0.54/0.78 \\
    Qwen3VL-8B (BB) & 0.30/0.29/0.31 & 0.84 & 0.44/0.60 \\
    Qwen3VL-8B (Dec.) & 0.55/0.54/0.54 & 0.95 & 0.52/0.81 \\
    AsyncDrive/nuPlan & 0.66 & 0.88 & 0.61/0.73 \\
    \bottomrule
    \end{tabular}
\end{minipage}

\vspace{-0.8em}
\end{figure*}

\begin{figure}[t]
\centering
\begin{minipage}[t]{0.42\linewidth}
    \centering
    \includegraphics[width=\linewidth,height=0.21\textheight,keepaspectratio]{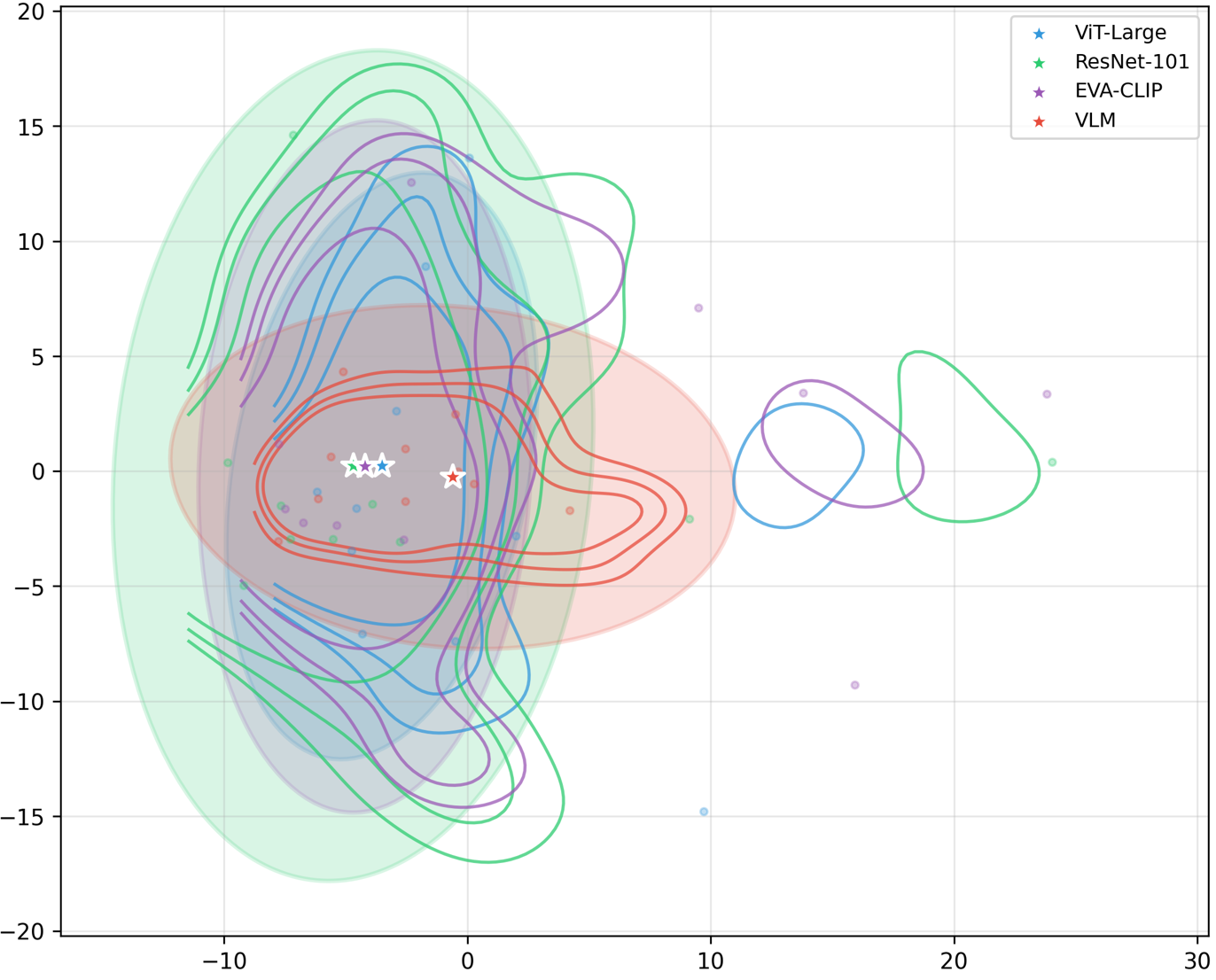}
    \captionof{figure}{
    \textbf{Aligned feature geometry after Procrustes alignment.}
    The VLM shares a common core with vision-only encoders while retaining model-specific residual regions.
    }
    \label{fig:aligned_feature_kde}
\end{minipage}
\hfill
\begin{minipage}[t]{0.56\linewidth}
    \centering
    \includegraphics[width=\linewidth,height=0.21\textheight,keepaspectratio]{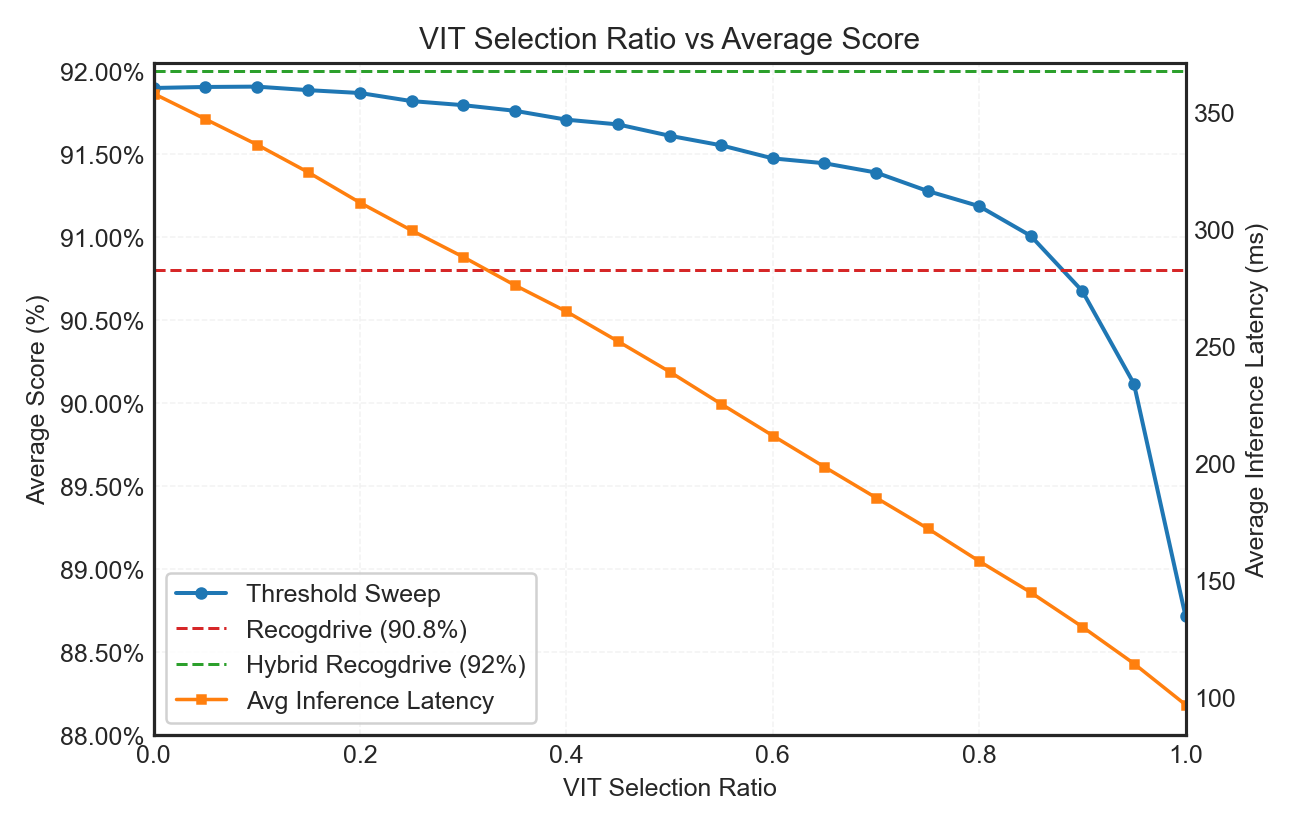}
    \captionof{figure}{
    \textbf{DualDriveVLA accuracy--compute trade-off.}
    Varying threshold $\gamma$ trades off fast-path ratio, PDMS, and inference latency.
    }
    \label{fig:rq3_tradeoff}
\end{minipage}

\vspace{-0.8em}
\end{figure}

\textbf{RQ1: representation.}
How similar are VLM and vision-only representations, and where do the differences remain? 
Using linear CKA, CCA, and a Shared--Unique SAE, we find that downstream policy learning substantially increases cross-branch alignment: in the main VLM--ViT setting, CKA rises from roughly $0.22$ at the backbone level to $0.54$ at the decision level. 
However, the branches do not collapse into a single redundant representation. 
SAE/CCA analyses consistently reveal a large shared decision subspace together with persistent non-transferable residual factors. 
Moreover, representation-only gates built from alignment or latent statistics provide only marginal gains over the VLM baseline (best PDMS $90.80 \rightarrow 90.96$), far below the oracle upper bound. 
This negative result motivates moving from static representation cues to trajectory- and behavior-level reasoning.

\textbf{RQ2: behavior.}
Do these residual factors translate into meaningful planning differences? 
We find stable but long-tailed complementarity rather than simple dominance. 
At the aggregate level, VLM and vision-only policies exhibit different progress--braking and path-choice tendencies. 
At the scenario level, vision-only encoders are relatively stronger in simple geometry-dominant scenes, while VLM-specific factors are more useful in semantically complex, ambiguous, and interaction-heavy cases. 
Under conservative counting, both branches decisively win on persistent subsets of scenarios, and simply choosing the better trajectory between a VLM branch and a ViT branch yields an oracle best-of-two score of $93.58$ PDMS on NAVSIM. 
Latent-intervention policies further support that the SAE-separated residual factors are behaviorally useful rather than merely reconstruction artifacts.

\textbf{RQ3: system.}
How can this complementarity be converted into practical accuracy--cost gains? 
We treat VLM and vision-only policies as complementary candidate generators rather than competitors to be averaged away. 
\emph{HybridDriveVLA} runs both branches, optionally augments their endpoints with simple interpolated candidates, and uses a learned trajectory scorer for selection, improving PDMS from $90.80$ to $92.10$ without changing policy training. 
\emph{DualDriveVLA} further turns the same idea into a fast--slow deployment policy: it runs the vision-only branch by default and invokes the VLM only when the fast-path score is insufficient. 
Calling the VLM in only about $15\%$ of scenarios achieves $91.00$ PDMS while reducing latency by about $1.9\times$ relative to the VLM baseline.

Overall, our contribution is an analysis-to-mechanism account of VLA driving: policy learning creates substantial shared structure between VLM and vision-only branches, but preserves behaviorally useful residual factors. 
Rather than distilling these differences away, we show that they can be exploited through lightweight trajectory-level selection and fast--slow deployment.

\section{Related Work}

\noindent\textbf{VLMs for autonomous driving.}
Recent work incorporates vision--language models into autonomous driving in two main ways. One line follows a \emph{dual-system} design, where a VLM provides high-level commands, semantic guidance, or intermediate waypoints for a downstream planner~\citep{tian2024drivevlm,jiang2024senna}. Another line adopts a more \emph{single-system} formulation, casting planning or action generation directly as multimodal language generation, often with prompting or textual reasoning for interpretability~\citep{wang2024omnidrive,mao2023language,bai20243d,shao2024lmdrive,zhang2024wisead,mao2023gpt,zhao2025sce2drivex,hwang2024emma,wei2022chain,fu2025orion}. We are closer in spirit to the dual-system view, but with a different emphasis: rather than treating the VLM mainly as a stronger semantic module, we ask how VLM and vision-only branches differ after policy learning, and how their residual complementarity can be converted into trajectory selection and efficient deployment.

\noindent\textbf{Diagnosing VLM/VLA driving stacks.}
Most existing analyses of VLM/VLA driving systems focus on interpretability, reasoning ability, or benchmark construction, such as explanation-centric datasets, language-grounded evaluation, and structured reasoning protocols~\citep{chonghao2023drivelm,ming2023reason2drive,tianshuai2025visionlanguageaction,sicong2025a,yue2025finegrained,zecong2026autodridm}. In contrast, we study a more mechanistic question inside a unified planning stack: how VLM and vision-only encoders differ in internal representations, how these differences are transformed---and often compressed---after downstream policy learning, and when the remaining residual factors become behaviorally useful at the scenario level. To our knowledge, this is among the first works to systematically connect \emph{representation geometry}, \emph{behavioral complementarity}, and \emph{trajectory-level selection} within a unified VLA-style planning framework.

Overall, our work complements prior VLM-for-driving systems by offering an analysis-to-mechanism pipeline: from shared-versus-unique representation structure (RQ1), to scenario-level behavioral complementarity (RQ2), to lightweight hybrid/dual designs that exploit this complementarity in practice (RQ3).

\section{Preliminaries}

\subsection{Formulation and Analysis Points}

In our NAVSIM experiments, a policy maps visual observation $I_{\mathrm{cam}}$, navigation signal $L_{\mathrm{nav}}$, and ego state $S_{\mathrm{ego}}$ to a future trajectory 
$\hat{\tau}=\Phi(I_{\mathrm{cam}},L_{\mathrm{nav}},S_{\mathrm{ego}})\in\mathbb{R}^{T\times 3}$, 
where each waypoint contains $(x_t,y_t,\theta_t)$.\footnote{The notation extends to multi-view inputs by letting $I_{\mathrm{cam}}=\{I^{(v)}\}_{v=1}^{V}$. We use a single front camera.}

For RQ1--RQ2, we probe two representations in Fig.~\ref{fig:main}: 
the \textbf{backbone feature} $\mathbf{h}^{\mathrm{bb}}$ after the encoder adapter, and the \textbf{decision feature} $\mathbf{h}^{\mathrm{dec}}$ immediately before the action head. 
The former isolates encoder-level geometry, while the latter captures what remains after downstream policy learning in the diffusion Transformer planner.

\subsection{Unified Two-branch Framework and Comparison Scope}

\begin{figure*}[t]
    \centering
    \includegraphics[width=1\linewidth]{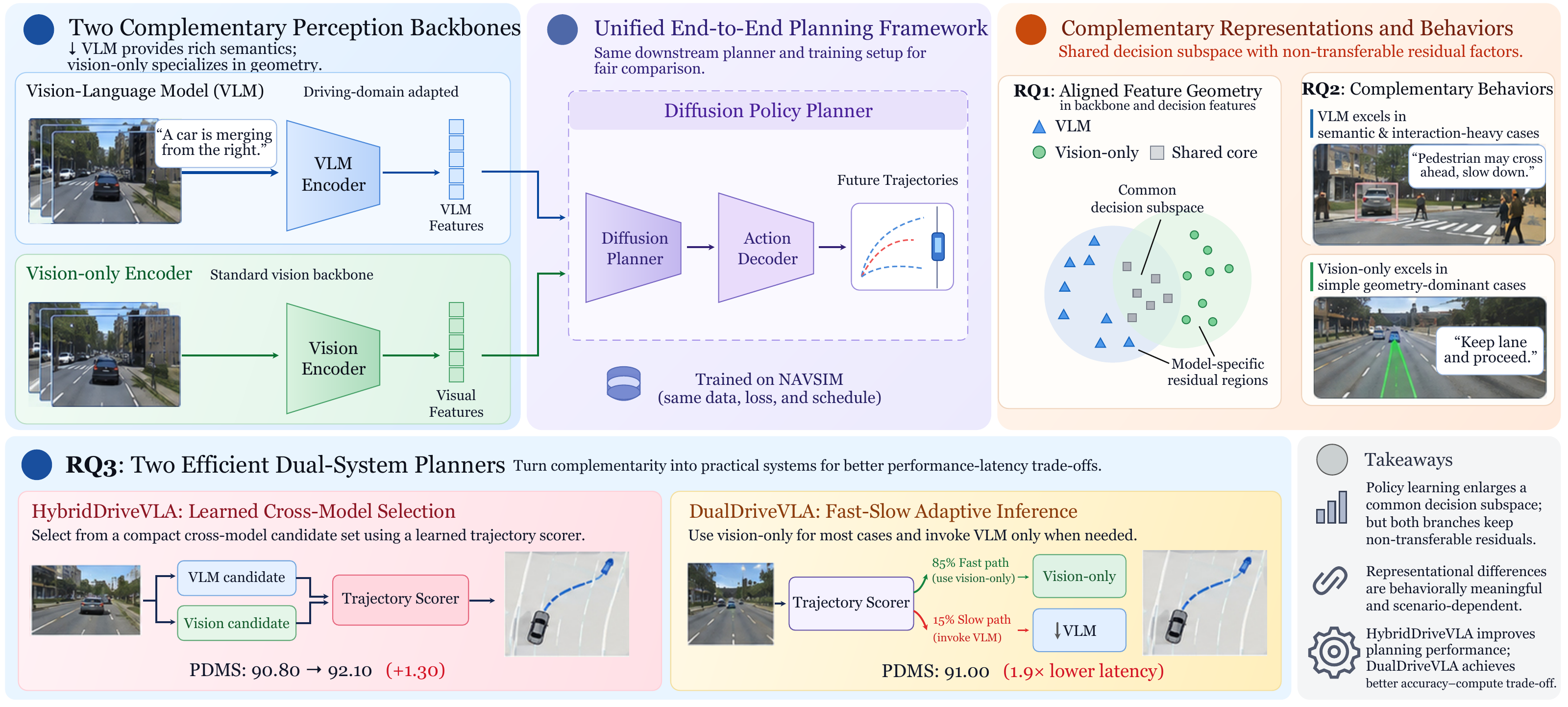}
    \caption{
    \textbf{Overview of our analysis-to-mechanism pipeline.}
    We compare a VLM branch and a vision-only branch under the same downstream planner/action architecture, probing both backbone and decision features.
    RQ1 analyzes shared versus branch-specific representation structure; RQ2 connects the residual factors to scenario-level behavioral complementarity; RQ3 exploits this complementarity through trajectory-level selection and fast--slow deployment.
    }
    \label{fig:main}
\end{figure*}

Figure~\ref{fig:main} illustrates our two-branch VLM-hidden + diffusion-policy framework. 
One branch uses a driving-domain-adapted VLM encoder as a frozen source of language-grounded visual tokens, following RecogDrive-style adaptation~\citep{li2025recogdrive}; the other replaces it with a standard vision-only encoder such as ViT, ResNet, or EVA-CLIP. 
Both branches project encoder outputs to the same planner width and use the same DiT planner, action head, trajectory format, data split, and training schedule, but are trained as separate policies without weight sharing.
During imitation learning, the driving-domain-adapted VLM encoder remains frozen, whereas the vision-only encoder is trained jointly with the planner and action head.
During subsequent reinforcement-learning refinement, both encoders are frozen and only the planner and action head are updated.

Our comparison is therefore between two practical representation pipelines under a controlled downstream planner, rather than a fully symmetric upstream pretraining-budget comparison. 
This distinction is important because current VLA driving systems commonly adapt a generic VLM to driving data before freezing it as a representation source, whereas vision-only E2E planners typically start from strong visual checkpoints and learn the driving policy through downstream planner training. 
Thus, differences should be attributed to the representation pipelines made available to the same planner, not to changes in the planner or action head. 
Full architectural and training details are provided in Appendix~\ref{app:recogdrive_details}.

\textbf{Domain-adaptation control.}
To test whether complementarity is mainly due to extra driving-domain exposure in the VLM branch, we add a NAVSIM-restricted control in Appendix~\ref{app:domain_adaptation_control}. 
\textbf{VLM-NAVSIM-DA} restricts VLM adaptation to NAVSIM training data and its available text/trajectory annotations. 
\textbf{ViT-NAVSIM-DA} initializes from the same public ViT-large checkpoint, performs MAE adaptation on NAVSIM training images only, discards the MAE decoder, and then follows the original ViT downstream planner training protocol. 
No NAVSIM test images or evaluation labels are used during adaptation. 
This control does not make VLM and ViT upstream objectives identical, but tests the narrower confound of whether the VLM--vision complementarity disappears once both branches receive in-domain adaptation on the same dataset.

\section{RQ1: Representation Analysis}

\subsection{Paired Features and Preprocessing}

We analyze paired representations extracted from the same driving scenarios. 
Let $(x_i,y_i)_{i=1}^n$ be aligned feature pairs from the VLM and vision-only branches, stacked as $X,Y\in\mathbb{R}^{n\times d}$. 
We probe two levels. 
At the \textbf{backbone level} ($d=384$), the VLM branch produces adapted tokens $F_{\mathrm{vlm}}\in\mathbb{R}^{L\times384}$, with $L=2800$ in the main setting; we use $x=\mathrm{MeanPool}(F_{\mathrm{vlm}})$ for tractability, while the vision-only branch provides an adapted global embedding $y\in\mathbb{R}^{384}$.\footnote{Non-pooled analyses on random subsets yield the same qualitative conclusions; details are provided in the appendix~\ref{app:Token-level}.}
At the \textbf{decision level} ($d=512$), we extract the planner representation after the final denoising step and immediately before the action head for both branches. 
Unless stated otherwise, features are centered across samples; SAE additionally uses training-split z-score normalization (Appendix~\ref{app:rq1_preprocess}).

\subsection{Linear Similarity: CKA and CCA}

We use two complementary linear tools. 
Linear CKA measures global geometric similarity, computed on centered features as 
$\mathrm{CKA}(X,Y)=\|X^\top Y\|_F^2/(\|X^\top X\|_F\|Y^\top Y\|_F)$, 
and is reported at both backbone and decision levels to test whether policy learning increases cross-branch alignment. 
PCA-whitened CCA measures alignability through maximally correlated linear directions. 
After PCA truncation and whitening for numerical stability, we obtain canonical correlations $\rho_1\ge\cdots\ge\rho_k$ and summarize them by the spectrum and aggregates such as mean@k; full details are in Appendix~\ref{app:cca_whiten}.

\subsection{Shared--Unique Sparse Autoencoder (SAE)}

CKA/CCA quantify similarity but do not separate shared from branch-specific factors, nor test whether the shared component is functionally interchangeable. 
We therefore train a Shared--Unique SAE on standardized paired features. 
For each pair $(x,y)$, the SAE encodes shared and unique latents $(z_s^x,z_u^x)=f^x(x)$ and $(z_s^y,z_u^y)=f^y(y)$, and reconstructs with additive linear decoders 
$\hat{x}=W_s^xz_s^x+W_u^xz_u^x+b^x$ and 
$\hat{y}=W_s^yz_s^y+W_u^yz_u^y+b^y$. 
This additive form lets us measure shared and unique contributions in the original feature space. 
Unless stated otherwise, we use $d_s=64$ and $d_u=16$.

Intuitively, the objective seeks a partially interchangeable shared component and compact branch-specific residuals.
Full and shared-only reconstruction preserve information; cross shared-only reconstruction (e.g., decoding $x$ from $z_s^y$) encourages partial interchangeability.
VICReg-style alignment, variance, and decorrelation terms align shared codes while discouraging collapse and redundancy~\citep{bardes2022vicreg}.
Shared--unique orthogonality discourages information leakage between components, and $\ell_1$ sparsity promotes compact residual codes.
These are established representation-factorization ingredients, with related precedents in Factorized Orthogonal Latent Spaces~\citep{salzmann2010factorized}, Domain Separation Networks~\citep{bousmalis2016domain}, Learning Factorized Multimodal Representations~\citep{tsai2019learning}, and MISA~\citep{hazarika2020misa}; the full objective is in Appendix~\ref{app:sae_full}.

For evaluation, we report explained variance $R^2$ for full, shared-only self, and shared-only cross reconstruction, and use the self--cross gap 
$\Delta_{\mathrm{cross}}=R^2_{\mathrm{sh}}-R^2_{\mathrm{cross}}$ 
as the main interchangeability statistic (Appendix~\ref{app:sae_metrics}).

\subsection{Results \& Discussion}
\label{sec:rq1_results}

\textbf{Policy learning increases alignment without causing collapse.}
We first compare the VLM and vision-only branches in the original feature space using linear CKA and PCA-whitened CCA.
In the main InternVL3-2B setting, VLM-to-vision CKA increases from about $0.22$ at the backbone level to about $0.54$ at the decision level (Fig.~\ref{fig:cka_backbone_vs_dit}), indicating stronger cross-branch alignment after the shared diffusion planner.
CCA shows the same trend from a subspace perspective: although mean canonical correlation changes only mildly ($0.61\!\rightarrow\!0.63$), the number of high-correlation directions with $\rho>0.8$ expands from $5/28$ to $28/78$, and the aligned energy increases from $28\%\!\rightarrow\!53\%$ for VLM and $56\%\!\rightarrow\!77\%$ for ViT.
Together with the aligned geometry in Fig.~\ref{fig:aligned_feature_kde}, this suggests that policy learning compresses heterogeneous encoder evidence into a larger shared decision space, but does not make the two branches redundant.

\textbf{Shared--Unique SAE as a substitutability diagnostic.}
CKA/CCA measure alignment, but not whether the shared factors inferred from one branch can substitute for those of the other.
We therefore use the Shared--Unique SAE to evaluate partial cross-branch substitutability through cross shared-only reconstruction $R^2_{\mathrm{cross}}$ and the self--cross gap $\Delta_{\mathrm{cross}}$.
Table~\ref{tab:sae_main} should be read as a structured diagnostic rather than causal evidence: shared-space alignment is encouraged by the SAE objective and is therefore treated as a sanity check, while our main interpretation relies on original-space CKA/CCA, $R^2_{\mathrm{cross}}$, and $\Delta_{\mathrm{cross}}$.
Decision-level features are more transferable than backbone features, especially on the ViT side, yet cross reconstruction remains far from perfect and the VLM-side improvement is modest.
Thus, downstream policy learning enlarges the shared component while preserving branch-specific residual factors.
Their behavioral relevance is tested later through scenario-level analysis and latent-intervention policies.

\begin{table*}[t]
  \caption{
  \textbf{SAE/CCA summary in the main setting.}
  $R^2_{\mathrm{cross}}$ and $\Delta_{\mathrm{cross}}$ measure partial substitutability of learned shared factors.
  Decision features are more aligned and partly more transferable, but cross reconstruction remains imperfect.
  }
  \label{tab:sae_main}
  \centering
  \fontsize{8pt}{9.5pt}\selectfont
  \setlength{\tabcolsep}{4pt}
  \renewcommand{\arraystretch}{0.95}
  \begin{tabular}{@{}lccccc@{}}
    \toprule
    Feature
    & CKA$_{\mathrm{orig}}$
    & $R^2_{\mathrm{cross}}(x{\leftarrow}z_s^y/y{\leftarrow}z_s^x)$
    & $\Delta_{\mathrm{cross}}(x/y)$
    & CCA mean@10
    & CCA AER
    \\
    \midrule
    Backbone & 0.213 & 0.537/0.623 & 0.098/0.160 & 0.800 & 0.286/0.556 \\
    Decision & 0.537 & 0.546/0.763 & 0.071/0.063 & 0.972 & 0.534/0.771 \\
    \bottomrule
  \end{tabular}
\end{table*}

\textbf{The shared-plus-unique pattern generalizes beyond the main setting.}
Table~\ref{tab:rq1_generalization} summarizes additional evidence across larger VLMs on NAVSIM and an AsyncDrive/nuPlan cross-stack check.
Across InternVL3-8B and Qwen3VL-8B, backbone alignment remains moderate, decision-level alignment increases after policy learning, and SAE cross reconstruction remains non-trivial.
The AsyncDrive/nuPlan result shows the same qualitative shared-plus-unique pattern under a different planner/dataset setting.
These results may not prove universality, but reduce the concern that the observed pattern is specific to one VLM family or planning stack.


\textbf{SAE sanity check.}
Because the SAE objective explicitly encourages shared-latent alignment, shared-space CKA often saturates and is used only as a training-validity check.
A shuffled-pair control, which randomly permutes $(x_i,y_i)$ pairings while keeping marginals fixed, substantially reduces shared-space alignment and collapses original-space alignment to near zero (Appendix~\ref{app:shuffle_control}).
We therefore base our main comparisons on original-space alignment and substitutability metrics, namely CKA/CCA, $R^2_{\mathrm{cross}}$, and $\Delta_{\mathrm{cross}}$.

\subsection{Negative Result: Representation-only Gating is Insufficient}
\label{sec:rq1_gating_negative}

Before turning to RQ2, we test whether one can select between the VLM and ViT trajectories using representation-level signals alone.
The gate uses backbone/decision features and SAE-derived shared/unique statistics to predict which branch will obtain the higher closed-loop score.
We evaluate handcrafted rules based on shared/unique energy ratios and learned gates including tree-based models, MLPs, and attention-based predictors (Appendix~\ref{app:gating_rules}, \ref{app:gating_models}).

Across settings, representation-only gating does not reliably outperform the VLM baseline and remains far below the oracle best-of-two (Table~\ref{tab:gating_main}).
Even the best learned gate improves PDMS only marginally, from $90.80$ to $90.96$, whereas the oracle reaches $93.58$.
Feature visualizations also show poor separation between ``VLM-better'' and ``ViT-better'' cases.
This negative result motivates RQ2: reliable selection requires trajectory- and behavior-level signals rather than static representation cues alone.

\subsection{Takeaways (RQ1)}

RQ1 shows that policy learning enlarges the shared VLM--vision subspace, but does not remove branch-specific residuals.
Shared--Unique SAE provides a substitutability test showing that decision features are only partially transferable, and the same shared-plus-unique pattern appears across larger VLMs and an extra planner/dataset setting.
However, representation-only gating yields only marginal gains over the VLM baseline, motivating the behavioral analysis in RQ2.

\section{RQ2: Behavioral Complementarity}
\label{sec:rq2}

\subsection{Scenario-level complementarity in the long tail}
\label{sec:rq2_scenario}

To characterize behavioral complementarity, we assign each scenario $i$ and policy $m$ an offline trajectory-quality score $s(m,i)$ (NAVSIM PDMS; Appendix~\ref{app:offline_score}). 
For a VLM--vision pair, we define the advantage as $\Delta_i(m)=s(\mathrm{VLM},i)-s(m,i)$ and count a significant win when $|\Delta_i(m)|>\tau$.

A recurring pattern is that complementarity is long-tailed: most scenarios show small score differences, but both branches decisively win on non-trivial and non-nested subsets. 
Using a stability-aware counting protocol (Appendix~\ref{app:rq2_win_count}), we observe 257 VLM wins vs.\ 253 ViT wins at $\tau=0.2$, and 159 vs.\ 153 at $\tau=0.5$. 
Thus, neither branch strictly contains the other on hard cases, motivating a candidate-generator view rather than a purely average-score comparison.

To identify where the long-tail differences arise, we further use $\tau=0.9$ to isolate 279 major-decision-mismatch cases and group them into six coarse scene types using the annotation protocol in Appendix~\ref{app:rq2_taxonomy}.
VLM wins 176 cases and ViT wins 103. 
The asymmetry becomes clear after semantic grouping: ViT wins more often only in simple lane keeping ($55/92$), whereas VLM dominates in intersection semantics ($58/78$), curve/merge/ramp cases ($23/38$), work zones/cones ($12/15$), narrow or unclear boundaries ($34/44$), and occlusion/dense clutter ($12/12$). 
Merged into simple vs.\ complex scenes, VLM wins $139/187$ ($74.3\%$) complex cases, while ViT is stronger in the simple subset.

This breakdown links RQ2 back to RQ1: VLM-specific residual factors are not merely abstract representation leftovers, but become useful in semantic disambiguation, boundary uncertainty, and interaction-heavy scenes. 
The occlusion/dense-clutter subset is especially indicative, where all 12 cases favor VLM with near-saturated scores while ViT collapses.

\subsection{Latent intervention: do unique factors have behavioral utility?}
\label{sec:latent_intervention}

We next test whether SAE-separated unique factors are functional rather than merely reconstructive artifacts. 
We do not claim strict causal identification, since SAE is not a full causal disentanglement method. 
Instead, after decomposing decision-level features into shared and branch-specific latents, we train policies with different latent subsets while keeping the downstream architecture and training protocol fixed.

Table~\ref{tab:latent_intervention} shows that shared-only latents already yield 83.0 PDMS, but adding either unique component improves performance. 
VLM-unique factors provide a larger overall gain than ViT-unique factors (86.9 vs.\ 86.1), and using both reaches 87.5. 
More importantly, in the 187 complex scenes, shared+VLM-unique reaches high scores in 86 cases, compared with 37 for shared+ViT-unique and 33 for shared-only; conversely, ViT-unique is more helpful in simple geometry-dominant scenes. 
Thus, while not a formal causal proof, this intervention-style evidence supports that unique factors have systematic, scene-conditioned behavioral utility.

\begin{table*}[t]
\centering
\begin{minipage}[t]{0.39\textwidth}
\centering
\caption{
\textbf{Latent-intervention study.}
Policies use different SAE-decomposed decision latents.
}
\label{tab:latent_intervention}
\small
\setlength{\tabcolsep}{4pt}
\renewcommand{\arraystretch}{0.92}
\begin{tabular}{@{}lcc@{}}
\toprule
Latents used & PDMS  & Complex  \\
\midrule
Shared only & 83.0 & 33 / 187 \\
Shared + ViT-unique & 86.1 & 37 / 187 \\
Shared + VLM-unique & 86.9 & 86 / 187 \\
Shared + both unique & 87.5 & 99 / 187 \\
\bottomrule
\end{tabular}
\end{minipage}
\hfill
\begin{minipage}[t]{0.58\textwidth}
\centering
\caption{
\textbf{Behavioral diagnostics on NAVSIM.}
Sig. denotes scenarios where the two branches differ substantially in score.
}
\label{tab:behavior_safety_main}
\small
\setlength{\tabcolsep}{4pt}
\renewcommand{\arraystretch}{0.92}
\begin{tabular}{@{}lcccc@{}}
\toprule
Metric & VLM & ViT & VLM sig. & ViT sig. \\
\midrule
Mean speed (m/s) & 5.31 & 5.17 & 5.74 & 5.43 \\
Path length (m) & 18.60 & 18.09 & 20.08 & 19.02 \\
Min long. acc. (m/s$^2$) & -2.87 & -2.69 & -2.99 & -2.53 \\
Cum. deceleration & 4.94 & 4.63 & 5.26 & 4.38 \\
Hard-brake segs. & 0.49 & 0.44 & 0.49 & 0.36 \\
\bottomrule
\end{tabular}
\end{minipage}
\vspace{-0.6em}
\end{table*}

\subsection{Behavior differences beyond scenario counts}
\label{sec:rq2_behavior}

Complementarity also appears in how the two branches trade off progress, braking, and path choice. 
Table~\ref{tab:behavior_safety_main} shows that, relative to ViT, the VLM branch has slightly higher progress-related quantities on the full test set, including mean speed and path length, and stronger braking-side responses, including more negative minimum longitudinal acceleration, larger cumulative deceleration, and more hard-brake segments. 
These gaps become larger on the significant-difference subset, suggesting that VLM tends to advance further while applying stronger corrections when needed, rather than simply driving uniformly faster.

Beyond longitudinal statistics, we also observe lateral differences in lane-centering, merge timing, and route selection. 
In many cases, the expert trajectory lies between the two branches or follows an intermediate path, motivating the interpolation-based candidate construction in Sec.~\ref{sec:rq3}. Appendix~\ref{app:behavior_safety} reports qualitative examples, the full behavioral and safety diagnostics with standard deviations over five random seeds, showing that the trends are stable across runs.

\subsection{Implications: selection needs trajectory-level signals}
\label{sec:rq2_implications}

If complementarity is real, selecting the better trajectory from a combined candidate set should yield a meaningful upper bound. 
For each scenario with candidate set $\mathcal{C}_i$, we define the Best-of-$n$ score as $S_{\mathrm{BoN}}(i)=\max_{\tau\in\mathcal{C}_i}s(\tau,i)$. 
With $\mathcal{C}_i=\{\tau_{\mathrm{VLM}}(i),\tau_{\mathrm{ViT}}(i)\}$, Best-of-2 improves PDMS from 90.80 to 93.58 (Table~\ref{tab:rq3_main}), confirming set-level complementarity.

Together with RQ1, these results show that reliable selection should rely on trajectory-/behavior-level signals rather than static representation cues alone. 
A practical implication is to score a small set of complementary candidates---the VLM and ViT trajectories and their interpolations---to recover part of the long-tail Best-of-$n$ gain without relying on large within-model sampling.

\section{RQ3: From Behavioral Complementarity to Trajectory Selection}
\label{sec:rq3}

RQ1 shows that representation-level cues are insufficient for reliable sample-wise selection, while RQ2 shows stable trajectory-level complementarity between VLM and vision-only policies. 
RQ3 therefore turns this complementarity into a practical mechanism: construct a small cross-model candidate set and select using trajectory-level scores. 
This is motivated by the limited diversity of within-model diffusion sampling, where increasing samples from a single ViT or VLM gives only marginal Best-of-$n$ gains (Table~\ref{tab:rq3_main}), consistent with prior observations for RecogDrive-style planners~\citep{chen2026devil}. 
We evaluate on NAVSIM \textit{navtest} using its pseudo closed-loop metrics: PDMS for NAVSIM-v1 and EPDMS for NAVSIM-v2 (Appendix~\ref{app:offline_score}). Tables~\ref{tab:comparison_modified} and~\ref{tab:exp_results} compare against prior and concurrent methods, complementing the ablations in Table~\ref{tab:rq3_main}.

\begin{table*}[t]
\centering
\begin{minipage}[t]{0.485\textwidth}
\captionsetup{type=table}
\caption{Representation-only gating on NAVSIM. Higher is better; oracle = per-scenario best branch.}
\label{tab:gating_main}
\centering
\fontsize{8pt}{9.5pt}\selectfont
\setlength{\tabcolsep}{4pt}
\renewcommand{\arraystretch}{0.95}
\begin{tabular}{@{}l c@{}}
\toprule
Method  & PDMS(\%)$\uparrow$ \\
\midrule
InternVL3-2B  & 90.80 \\
ViT-base/large & 85.62/88.88 \\
Resnet-50/101 & 86.02/87.69 \\
Evaclip02-base & 87.89 \\
Oracle best-of-two (VLM+ViT-large) & \textbf{93.58} \\
\midrule
Rule: More-unique wins & 89.95 \\
Rule: Shared-dominant conditional & 90.22 \\
Rule: Smoothed shared-dominance & \textbf{90.29} \\
Rule: ViT-prior fallback & 89.92 \\
\midrule
Random Forest & 90.87 \\
Gradient Boosting & 90.75 \\
GBDT & 90.65 \\
MLP classifier & 90.80 \\
Self-attention (binary/regression) & 90.82/\textbf{90.96} \\
Self-attention (partial score terms) & 90.82 \\
\bottomrule
\end{tabular}
\end{minipage}\hfill
\begin{minipage}[t]{0.485\textwidth}
\captionsetup{type=table}
\caption{
\textbf{Trajectory-level selection and candidate augmentation on NAVSIM (\textit{navtest}).}
Endpoint-only scorer selection already captures much of the cross-model complementarity; interpolation provides an additional lightweight candidate-set augmentation.
DualDriveVLA uses the fast branch by default (Sec.~\ref{sec:rq3_dual}).
}
\label{tab:rq3_main}
\centering
\fontsize{8pt}{9.5pt}\selectfont
\setlength{\tabcolsep}{4pt}
\renewcommand{\arraystretch}{0.95}
\begin{tabular}{@{}l c@{}}
\toprule
Method & PDMS(\%)$\uparrow$ \\
\midrule
InternVL3-2B  & 90.80 \\
ViT-large & 88.88 \\
\midrule
Best-of-$n$ (ViT, $n{=}1/3/6$) & 88.88/89.13/89.32  \\
Best-of-$n$ (VLM, $n{=}1/3/6$) & 90.80/91.57/91.95 \\
Cross-model oracle (Best-of-2) & 93.58 \\
Cross-model oracle (Best-of-6) & 94.00 \\
\midrule
Trajectory mean (two endpoints) & 91.18 \\
Rule-based selection (grid search) & 91.21 \\
Adaptive weighting (predict $\alpha$) & 91.31 \\
Scorer selection (endpoints only) & 91.75 \\
HybridDriveVLA & \textbf{92.10} \\

\bottomrule
\end{tabular}
\end{minipage}
\end{table*}

\subsection{HybridDriveVLA: cross-model candidates and trajectory scoring}
\label{sec:rq3_hybrid}

For each scenario, HybridDriveVLA starts from the two endpoint trajectories $\tau_{\mathrm{vit}}$ and $\tau_{\mathrm{vlm}}$. 
Motivated by RQ2, where the two branches often differ in progress--braking--path-choice trade-offs and expert-like trajectories can lie between them, we augment the endpoints with a simple interpolation set: 
$\tau_{\alpha}=\alpha\tau_{\mathrm{vit}}+(1-\alpha)\tau_{\mathrm{vlm}}$, with $\alpha\in\{0.1,\ldots,0.9\}$. 
The final candidate set is $\mathcal{C}=\{\tau_{\mathrm{vit}},\tau_{\mathrm{vlm}},\tau_{0.1},\ldots,\tau_{0.9}\}$. 
This interpolation is deliberately modest: it is not a general trajectory generator, but a low-cost local expansion around two complementary endpoints. 
Table~\ref{tab:rq3_main} shows that endpoint-only scorer selection already captures much of the complementarity, while interpolation provides an additional gain.

Given $\mathcal{C}$, we train a lightweight trajectory scorer to predict PDMS-related sub-score components, following the spirit of DrivoR-style scoring~\citep{kirby2026driving}. 
The scorer encodes the current scene with DINOv2-Small and embeds each candidate trajectory $\tau$ into a query $q_{\tau}=f_{\mathrm{wp}}(\tau)\in\mathbb{R}^{D_{\mathrm{score}}}$. 
Trajectory-conditioned heads then predict safety-, progress-, and comfort-related components, supervised by the oracle evaluator with component-wise losses. 
At inference, the predicted components are composed into a meta-score $\hat{s}(\tau)$ following the PDMS structure, and the selected trajectory is $\tau^\star=\arg\max_{\tau\in\mathcal{C}}\hat{s}(\tau)$. 
The scorer does not generate trajectories; it only ranks candidates from the complementary VLM and vision-only branches.

\begin{table*}[t!]
    \centering
    \small
    \caption{\textbf{Performance comparison on NAVSIM-v1 \textit{navtest} using PDMS.}}
    \label{tab:comparison_modified}
    \scalebox{0.85}{
    \begin{tabular}{@{}l|cc|ccc|c@{}}
        \toprule
        Method & NC$\uparrow$ & DAC$\uparrow$ & TTC$\uparrow$ & Comf.$\uparrow$ & EP$\uparrow$ & PDMS$\uparrow$ \\
        \midrule
        \multicolumn{7}{@{}l}{\textit{Traditional E2E planners}} \\
        \midrule
        UniAD~\citep{hu2023planning} & 97.8 & 91.9 & 92.9 & \textbf{100} & 78.8 & \cellcolor{gray!30} 83.4 \\

        Hydra-MDP~\citep{li2024hydra} & 98.3 & 96.0 & 94.6 & \textbf{100} & 78.7 & \cellcolor{gray!30} 86.5 \\
        
        DiffusionDrive~\citep{liao2024diffusiondrive} & 98.2 & 96.2 & 94.7 & \textbf{100} & 82.2 & \cellcolor{gray!30} 88.1 \\
        WoTE~\citep{li2025end} & 98.5 & 96.8 & 94.9 & 99.9 & 81.9 & \cellcolor{gray!30} 88.3 \\
        
        DiffusionDriveV2~\citep{zou2025diffusiondrivev2} & 98.3 & 97.9 & 94.8 & 99.9 & \textbf{88.0} & \cellcolor{gray!30} 91.2 \\
        iPad~\citep{guo2025ipad} & 98.6 & 98.3 & 94.9 & \textbf{100} & \textbf{88.0} & \cellcolor{gray!30} 91.7 \\
        \midrule
        \multicolumn{7}{@{}l}{\textit{VLA-based planners}} \\
        \midrule
        DrivingGPT~\citep{chen2024drivinggpt} & 98.9 & 90.7 & 94.9 & 95.6 & 79.7 & \cellcolor{gray!30} 82.4 \\
        PARA-Drive~\citep{weng2024drive} & 97.9 & 92.4 & 93.0 & 99.8 & 79.3 & \cellcolor{gray!30} 84.0 \\
        DRAMA~\citep{yuan2024drama} & 98.0 & 93.1 & 94.8 & 100 & 80.1 & \cellcolor{gray!30} 85.5 \\
        ImagiDrive~\citep{li2025imagidrive} & 98.1 & 96.2 & 94.5 & 100 & 80.5 & \cellcolor{gray!30} 86.9 \\
        AutoVLA~\citep{li2025drivevla} & 98.4 & 95.6 & \textbf{98.0} & 99.9 & 81.9 & \cellcolor{gray!30} 89.1 \\
        DriveVLA-W0~\citep{zhou2025autovla} & 98.7 & 99.1 & 95.3 & 99.3 & 83.3 & \cellcolor{gray!30} 90.2 \\
        Curious-VLA~\citep{chen2026devil} & 98.4 & 96.9 & 97.9 & 98.1 & 88.5 & \cellcolor{gray!30} 90.2 \\
        ReCogDrive~\citep{li2025recogdrive} & 97.9 & 97.3 & 94.9 & \textbf{100} & 87.3 & \cellcolor{gray!30} 90.8 \\
        WAM-diff~\citep{xu2025wam} & \textbf{99.1} & 98.3 & 96.5 & 99.9 & 84.4 & \cellcolor{gray!30} 91.0 \\
        \textit{HybridDriveVLA(ours)} & 98.6 & \textbf{98.6} & 96.2 & \textbf{100} & 87.3 & \cellcolor{gray!30} \textbf{92.1} \\
        \bottomrule
    \end{tabular}
    }
\end{table*}

\begin{table*}[ht]
  \centering
  \caption{\textbf{Performance comparison on NAVSIM-v2 \textit{navtest} using EPDMS.}}
  \label{tab:exp_results}
  \small
  \begin{adjustbox}{max width=\linewidth}
\scalebox{0.85}{
  \begin{tabular}{l|ccccccccc|c}
    \toprule
    Method              & NC$\uparrow$ & DAC$\uparrow$ & EP$\uparrow$ & TTC$\uparrow$ & C$\uparrow$ & TL$\uparrow$ & DDC$\uparrow$ & LK$\uparrow$ & EC$\uparrow$ & EPDMS$\uparrow$ \\
    \midrule
    Transfuser~\citep{chitta2022transfuser}    & 97.7 & 92.8 & 79.2 & 92.8 & \textbf{100}   & 99.9 & 98.3 & 67.6 & 95.3 & 77.8 \cellcolor{gray!30}\\
    VADv2~\citep{chen2024vadv2}         & 97.3 & 91.7 & 77.6 & 92.7 & \textbf{100}   & 99.9 & 98.2 & 66.0 & 97.4 & 76.6\cellcolor{gray!30} \\
    Hydra-MDP++~\citep{li2024hydra}   & 97.9 & 96.5 & 79.2 & 93.4 & \textbf{100}   & \textbf{100.0} & 98.9 & 67.2 & 97.7 & 80.6 \cellcolor{gray!30}\\
    ARTEMIS~\citep{feng2025artemis}           & 98.3 & 95.1 & 81.5 & 97.4 & \textbf{100} & 99.8 & 98.6 & 96.5 & \textbf{98.3} & 83.1 \cellcolor{gray!30}\\
    ReCogDrive~\citep{li2025recogdrive}  & 98.3 & 95.2 & 87.1 & 97.5 & 98.3 & 99.8 & \textbf{99.5} & \textbf{96.6} & 86.5 & 83.6 \cellcolor{gray!30}\\
    DiffusionDriveV2~\citep{zou2025diffusiondrivev2} & 97.7 & \textbf{96.6} & 88.9 & 97.2 & 97.8 & 99.8 & 99.2 & 96.0 & 91.0  & 85.5 \cellcolor{gray!30}\\
    \midrule
    \textit{HybridDriveVLA(ours)}  & \textbf{98.6} & 92.2 & \textbf{89.7} & \textbf{98.5} & 98.3 & 99.8 & 99.3 & \textbf{96.6} & 87.0 & \textbf{85.5} \cellcolor{gray!30}\\
    \bottomrule
  \end{tabular}}
\end{adjustbox}
  \vspace{-0.1in}
\end{table*}

\subsection{DualDriveVLA: fast--slow deployment}
\label{sec:rq3_dual}

HybridDriveVLA improves accuracy by always running both branches. 
DualDriveVLA converts the same complementarity into a fast--slow policy: it first runs the ViT branch and scores $\tau_{\mathrm{vit}}$; if $\hat{s}(\tau_{\mathrm{vit}})\ge\gamma$, it directly outputs the fast-path trajectory, otherwise it invokes the VLM branch, constructs the hybrid candidate set, and applies the scorer selection. 
Sweeping $\gamma$ exposes an accuracy--compute trade-off (Fig.~\ref{fig:rq3_tradeoff}): lower thresholds favor the fast path, while higher thresholds approach HybridDriveVLA by invoking the VLM more often. On a single NVIDIA A30 GPU, the VLM baseline obtains 90.80 PDMS with about 280 ms latency. With the recommended threshold, DualDriveVLA invokes the VLM in only about 15\% of scenarios, achieves 91.00 PDMS, and reduces latency to about 150 ms, corresponding to a $1.9\times$ speedup while slightly improving accuracy.

\section{Limitations}
Our analysis focuses on VLA driving systems that use a native VLM encoder as a source of language-grounded visual hidden states for an end-to-end planner. This enables comparison with vision-only encoders under the same downstream planner, isolating the representation and behavior differences introduced by the encoder pipeline.

The analysis is less directly applicable to LLM-centric architectures such as Orion~\citep{fu2025orion} or SimLingo~\citep{renz2025simlingo}, which inject tokens from an external visual front-end into a primarily text-based LLM. Without a standalone native VLM encoder, applying our shared--unique analysis requires redefining the comparison unit, for example by comparing visual front-ends before LLM fusion or probing representations before and after multimodal token fusion.

\section{Conclusion}
We studied whether VLM and vision-only backbones remain redundant or complementary after policy learning in end-to-end driving. 
Across representation, behavior, and system analyses, we find that downstream policy learning enlarges a shared decision subspace but preserves branch-specific residual factors that lead to long-tailed behavioral complementarity. 
Static representation-based gates cannot reliably exploit this complementarity, whereas trajectory-level selection over a small cross-model candidate set yields stable gains. 
HybridDriveVLA improves NAVSIM PDMS from 90.80 to 92.10, and DualDriveVLA retains 91.00 PDMS while invoking the VLM in only 15\% of scenarios, reducing latency from about 280 ms to 150 ms on a single NVIDIA A30 GPU. 

\bibliography{example_paper}
\bibliographystyle{plainnat}

\newpage
\appendix
\onecolumn



\appendix

\section{Detailed Formulas for RQ1}
\label{app:rq1_formulas}

\setcounter{equation}{0}

This appendix specifies the full definitions and implementation details omitted from the main text of RQ1, including preprocessing, linear CKA/CCA (with PCA truncation and whitening), and the Shared--Unique SAE objective and metrics.

\subsection{RecogDrive Details and Training Protocol}
\label{app:recogdrive_details}

\subsubsection{Architecture and Tensor Shapes}
\paragraph{VLM branch (InternVL3-2B).}
Given $I_{\mathrm{cam}}$ and a fixed textual prompt, the VLM produces last-layer hidden states
\[
H_{\mathrm{vlm}}\in\mathbb{R}^{L\times d_{\mathrm{vlm}}}.
\]
A linear adapter maps tokens to the planner width $d$:
\[
F_{\mathrm{vlm}} = H_{\mathrm{vlm}}W_{\mathrm{vlm}},\qquad
F_{\mathrm{vlm}}\in\mathbb{R}^{L\times d}.
\]
For feature analysis, the backbone feature is defined by mean pooling:
\[
\mathbf{h}^{\mathrm{bb}}_{\mathrm{vlm}}=\mathrm{MeanPool}(F_{\mathrm{vlm}})\in\mathbb{R}^{d}.
\]

\paragraph{Vision-only branch (ViT/ResNet/EVA-CLIP).}
Each vision-only backbone outputs a global embedding
\[
\tilde{\mathbf{h}}^{\mathrm{bb}}_{\mathrm{vis}}\in\mathbb{R}^{d_{\mathrm{vis}}},
\]
which is mapped to the same planner width:
\[
\mathbf{h}^{\mathrm{bb}}_{\mathrm{vis}}=\tilde{\mathbf{h}}^{\mathrm{bb}}_{\mathrm{vis}}W_{\mathrm{vis}}\in\mathbb{R}^{d}.
\]
For interface consistency, $\mathbf{h}^{\mathrm{bb}}_{\mathrm{vis}}$ can be implemented as a length-1 token sequence when the planner expects token inputs.

\paragraph{Diffusion planner and decision feature.}
A diffusion Transformer planner (DiT) iteratively denoises to produce a latent trajectory representation. The \emph{decision feature} is the planner output immediately before the action head:
\[
\mathbf{h}^{\mathrm{dec}}\in\mathbb{R}^{d_{\mathrm{dec}}}.
\]
A lightweight MLP action head maps it to $\hat{\tau}\in\mathbb{R}^{T\times 3}$.

\paragraph{Default dimensions.}
Unless stated otherwise, we use $d{=}384$, $d_{\mathrm{dec}}{=}512$, and $T{=}8$.

\subsubsection{Training Protocol and Fairness Controls}
All variants share the same NAVSIM configuration (split, optimizer, schedule, epochs) and identical planner/action-head architecture.

\paragraph{VLM branch.}
We follow the RecogDrive recipe:
(1) domain adaptation of the VLM on refined image--text driving data with trajectory supervision;
(2) imitation learning on NAVSIM with the VLM backbone frozen and the planner/action head trained;
(3) reinforcement learning on NAVSIM with the VLM still frozen and only the planner/action head updated.

\paragraph{Vision-only branch.}
Vision-only backbones are initialized from public pretrained weights (without the image--text domain-adaptation stage). On NAVSIM, we train the vision backbone jointly with the planner/action head during imitation learning. During RL refinement, the vision backbone is frozen and only the planner/action head is updated, matching the VLM branch to isolate backbone effects.

\subsubsection{Compute resources.}
All main training experiments were conducted on a server with $8$ NVIDIA A100 GPUs.
The VLM driving-domain adaptation stage takes approximately $2$--$3$ days depending on the VLM scale, with the $8$B variants being more expensive than the $3$B variants.
After obtaining the adapted/frozen encoder, the downstream planner training requires about $20$ hours for the imitation-learning stage and about $20$ hours for the subsequent reinforcement/refinement stage.
The reported latency measurements are obtained on a single NVIDIA A30 GPU, as described in Sec.~\ref{sec:rq3}.

\subsection{Paired Features, Centering, and Standardization}
\label{app:rq1_preprocess}

\paragraph{Paired features.}
Let $(x_i,y_i)_{i=1}^n$ be aligned feature pairs extracted from the same driving scenarios, and stack them as
\begin{equation}
X = [x_1^\top;\ldots;x_n^\top]\in\mathbb{R}^{n\times d},\qquad
Y = [y_1^\top;\ldots;y_n^\top]\in\mathbb{R}^{n\times d}.
\end{equation}
At the backbone level, $x_i$ is obtained by mean-pooling the VLM token sequence after the RecogDrive adapter (to support dataset-level statistics and cross-backbone comparability), while $y_i$ is the vision-only global embedding after its adapter (no pooling).

\paragraph{Centering across samples.}
We center each feature dimension across samples using
\begin{equation}
H = I - \frac{1}{n}\mathbf{1}\mathbf{1}^\top \in\mathbb{R}^{n\times n},
\qquad
\tilde{X} = HX,\ \tilde{Y} = HY.
\end{equation}
Unless stated otherwise, CKA/CCA use centered features.

\paragraph{Per-dimension z-scoring (used for SAE and reported $R^2$).}
For SAE training and all reported $R^2$ values, we standardize each dimension using \emph{training-split, dataset-level} statistics:
\begin{equation}
x' = (x-\mu_x)\oslash\sigma_x,\qquad
y' = (y-\mu_y)\oslash\sigma_y,
\end{equation}
where $\mu_x,\sigma_x\in\mathbb{R}^{d}$ are computed per-dimension over the training split (and then reused for validation/test).
For notational simplicity we drop primes in the SAE sections.
(For some visualizations only, we may apply plot-specific normalization; such choices do not affect the quantitative RQ1 results.)

\subsection{Linear Similarity: Linear CKA}
\label{app:cka}

Given centered matrices $\tilde{X},\tilde{Y}$, linear CKA is
\begin{equation}
\mathrm{CKA}(X,Y)
=
\frac{\|\tilde{X}^\top \tilde{Y}\|_F^2}
{\|\tilde{X}^\top \tilde{X}\|_F\;\|\tilde{Y}^\top \tilde{Y}\|_F}.
\end{equation}

\subsection{PCA-Truncation and Whitening for CCA}
\label{app:cca_whiten}

Given centered feature matrices $\tilde{X}\in\mathbb{R}^{n\times d_x}$ and $\tilde{Y}\in\mathbb{R}^{n\times d_y}$, define sample covariances
\begin{align}
\Sigma_{xx} &= \frac{1}{n-1}\tilde{X}^\top\tilde{X},\qquad
\Sigma_{yy} = \frac{1}{n-1}\tilde{Y}^\top\tilde{Y},\qquad
\Sigma_{xy} = \frac{1}{n-1}\tilde{X}^\top\tilde{Y}.
\end{align}

\paragraph{PCA truncation (explained-variance threshold).}
We eigendecompose
\begin{align}
\Sigma_{xx} = P_x \Lambda_x P_x^\top,\qquad
\Sigma_{yy} = P_y \Lambda_y P_y^\top,
\end{align}
where $\Lambda_x=\mathrm{diag}(\lambda^x_1,\ldots,\lambda^x_{d_x})$ and $\lambda^x_1\ge\cdots\ge 0$ (similarly for $y$).
We choose the smallest $k_x$ (resp.\ $k_y$) such that the cumulative explained variance reaches $\eta=0.99$:
\begin{equation}
k_x=\min\left\{k:\frac{\sum_{j=1}^{k}\lambda^x_j}{\sum_{j=1}^{d_x}\lambda^x_j}\ge \eta\right\},\qquad
k_y=\min\left\{k:\frac{\sum_{j=1}^{k}\lambda^y_j}{\sum_{j=1}^{d_y}\lambda^y_j}\ge \eta\right\}.
\end{equation}
We then keep
\begin{equation}
P_x^{(k_x)} = P_x[:,1{:}k_x],\ \Lambda_x^{(k_x)}=\Lambda_x[1{:}k_x,1{:}k_x],\qquad
P_y^{(k_y)} = P_y[:,1{:}k_y],\ \Lambda_y^{(k_y)}=\Lambda_y[1{:}k_y,1{:}k_y].
\end{equation}

\paragraph{Whitening (ridge-stabilized).}
We use ridge $\epsilon=10^{-8}$:
\begin{equation}
W_x = P_x^{(k_x)}\left(\Lambda_x^{(k_x)}+\epsilon I\right)^{-1/2},\qquad
W_y = P_y^{(k_y)}\left(\Lambda_y^{(k_y)}+\epsilon I\right)^{-1/2},
\end{equation}
and compute whitened features
\begin{equation}
\hat{X}=\tilde{X}W_x,\qquad \hat{Y}=\tilde{Y}W_y.
\end{equation}

\paragraph{Canonical correlations.}
CCA is obtained via SVD:
\begin{equation}
\hat{X}^\top \hat{Y} = U\,\mathrm{diag}(\rho_1,\dots,\rho_k)\,V^\top,
\end{equation}
where $k=\min(k_x,k_y)$ and $\rho_j\in[0,1]$ are canonical correlations.


\subsection{CCA Canonical-correlation Spectra and Original-space Aligned Energy}
\label{app:cca_spectrum}

We visualize the PCA-whitened CCA canonical-correlation spectra and report how much \emph{original-space} feature energy lies in highly aligned CCA directions.
Importantly, the ``CCA-aligned subspace'' here is a post-hoc linear construct and should not be conflated with the learned \emph{shared/unique} factors of the SAE.

\begin{figure}[t]
  \centering
  \begin{subfigure}{0.6\linewidth}
    \centering
    \includegraphics[width=\linewidth]{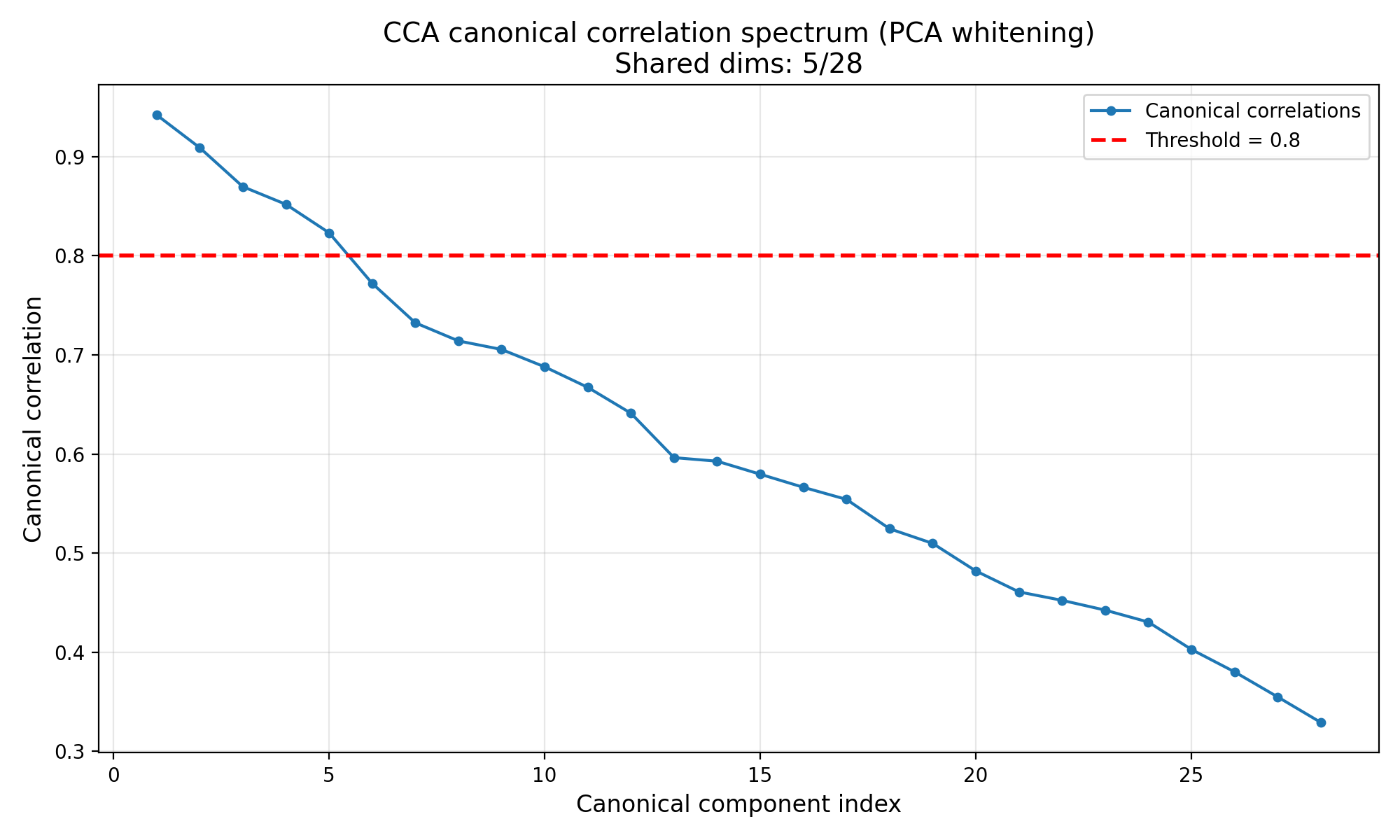}
    \caption{Backbone (28 dims).}
  \end{subfigure}
  \begin{subfigure}{0.6\linewidth}
    \centering
    \includegraphics[width=\linewidth]{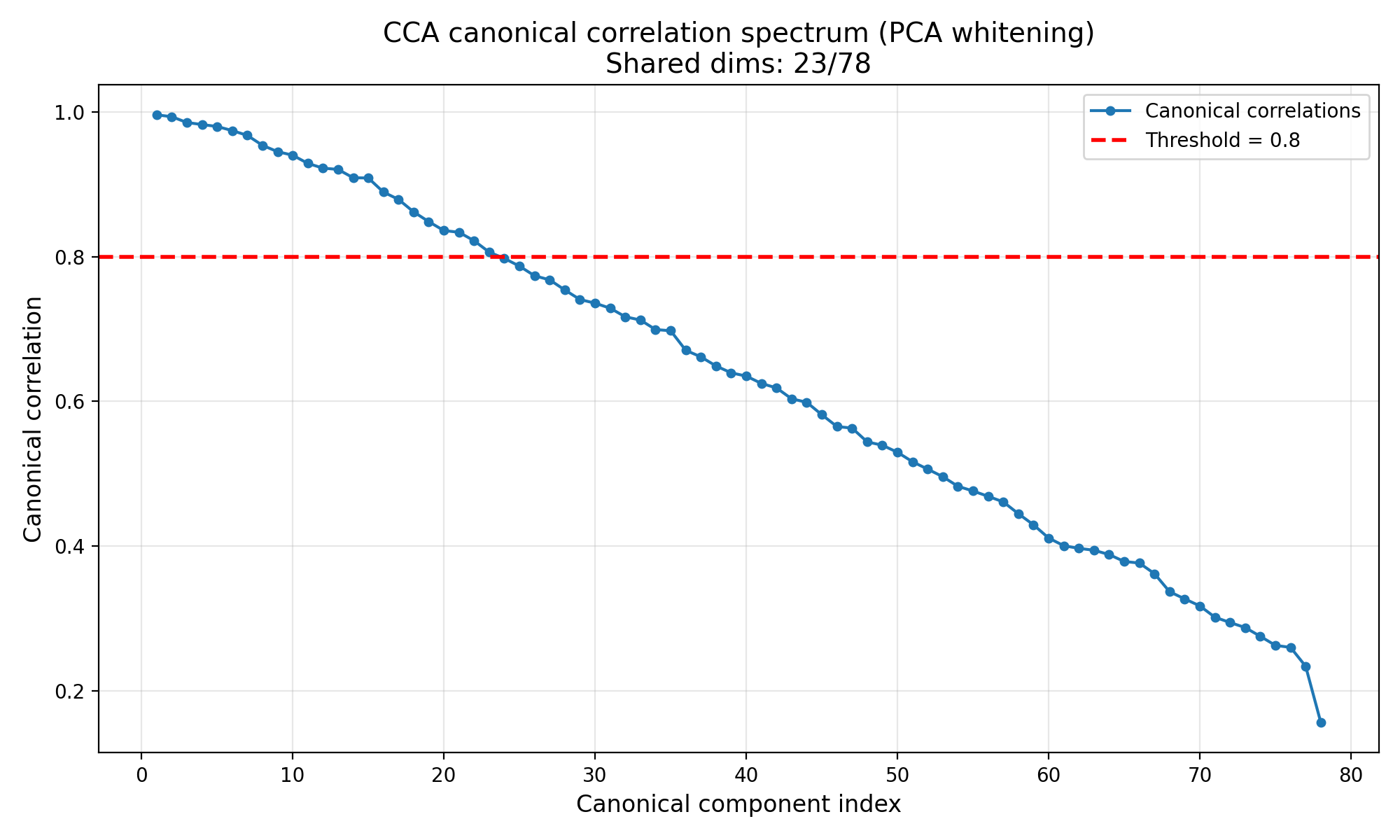}
    \caption{DiT (78 dims).}
  \end{subfigure}
  \caption{CCA canonical-correlation spectra (PCA truncation + whitening.}
  \label{fig:cca_spectrum}
\end{figure}

\paragraph{Canonical-correlation spectra.}
Figure~\ref{fig:cca_spectrum} shows the canonical correlations $\{\rho_j\}$ for backbone-level features (28 PCA-whitened dimensions) and DiT-level features (78 PCA-whitened dimensions). DiT yields a much larger set of near-perfectly aligned directions, consistent with increased decision-level isomorphism.

\paragraph{Original-space aligned-subspace energy (thresholded by $\rho>\tau$).}
Let $X\in\mathbb{R}^{n\times d_x}$ and $Y\in\mathbb{R}^{n\times d_y}$ be centered features.
After PCA truncation and whitening (Appendix~\ref{app:cca_whiten}), we obtain whitened features
$\tilde{X}=X U_X \Lambda_X^{-\frac12}$ and $\tilde{Y}=Y U_Y \Lambda_Y^{-\frac12}$.
Running CCA on $(\tilde{X},\tilde{Y})$ yields canonical directions $A,B$ and canonical correlations
$\rho_1\ge\cdots\ge\rho_k$.

For a threshold $\tau$ (we use $\tau=0.8$), define the index set
\[
\mathcal{I}_\tau=\{j:\rho_j>\tau\}.
\]
Map the selected canonical directions back to the original feature coordinates:
\[
Q_X = U_X \Lambda_X^{-\frac12} A_{\mathcal{I}_\tau},
\qquad
Q_Y = U_Y \Lambda_Y^{-\frac12} B_{\mathcal{I}_\tau}.
\]
Let $\mathrm{orth}(\cdot)$ return an orthonormal basis for the column span (e.g., via QR), and define
$\bar{Q}_X=\mathrm{orth}(Q_X)$ and $\bar{Q}_Y=\mathrm{orth}(Q_Y)$.
We then measure the fraction of \emph{original-space} energy captured by the CCA-aligned subspace as
\[
E_X^{\mathrm{full}}=\frac{1}{n}\|X\|_F^2,\quad
E_X^{\mathrm{align}}(\tau)=\frac{1}{n}\|X\bar{Q}_X\|_F^2,\quad
\mathrm{Frac}_X^{\mathrm{align}}(\tau)=\frac{E_X^{\mathrm{align}}(\tau)}{E_X^{\mathrm{full}}},
\]
and analogously for $Y$ using $\bar{Q}_Y$.

\paragraph{Numerical summary (used in Table~\ref{tab:sae_main}).}
Backbone level (28 dims): top-10 $\rho$ are
0.94, 0.91, 0.87, 0.85, 0.82, 0.77, 0.73, 0.71, 0.71, 0.69 (mean@10 $=0.800$).
With $\tau=0.8$, the aligned-energy fractions are $28.6\%$ (VLM) and $55.6\%$ (ViT).

DiT level (78 dims): top-10 $\rho$ are
0.996, 0.994, 0.986, 0.983, 0.980, 0.975, 0.968, 0.954, 0.946, 0.941 (mean@10 $\approx0.972$).
With $\tau=0.8$, the aligned-energy fractions are $53.4\%$ (VLM) and $77.1\%$ (ViT).

\subsection{Shared--Unique SAE: Full Objective and Regularizers}
\label{app:sae_full}

This section specifies the Shared--Unique SAE used in RQ1. SAE is trained in standardized feature space (Appendix~\ref{app:rq1_preprocess}).

\subsubsection{Encoders and additive linear decoders}
For a minibatch $\{(x^{(i)},y^{(i)})\}_{i=1}^{B}$,
\begin{align}
z_s^x &= f_s^x(x)\in\mathbb{R}^{B\times d_s}, & z_u^x &= f_u^x(x)\in\mathbb{R}^{B\times d_u},\\
z_s^y &= f_s^y(y)\in\mathbb{R}^{B\times d_s}, & z_u^y &= f_u^y(y)\in\mathbb{R}^{B\times d_u}.
\end{align}
We use MLP encoders (ReLU) and additive linear decoders
\begin{align}
\hat{x} &= W_s^x z_s^x + W_u^x z_u^x + \mathbf{1}(b^x)^\top,\\
\hat{y} &= W_s^y z_s^y + W_u^y z_u^y + \mathbf{1}(b^y)^\top,
\end{align}
where $W_s^x\in\mathbb{R}^{d\times d_s}$, $W_u^x\in\mathbb{R}^{d\times d_u}$ (and similarly for $y$), and $b^x,b^y\in\mathbb{R}^{d}$.

\paragraph{Shared-only reconstructions (self).}
\begin{align}
\hat{x}_s &= W_s^x z_s^x + \mathbf{1}(b^x)^\top,\qquad
\hat{y}_s = W_s^y z_s^y + \mathbf{1}(b^y)^\top.
\end{align}

\paragraph{Shared-only reconstructions (cross).}
\begin{align}
\hat{x}_{s\leftarrow y} &= W_s^x z_s^y + \mathbf{1}(b^x)^\top,\qquad
\hat{y}_{s\leftarrow x} = W_s^y z_s^x + \mathbf{1}(b^y)^\top.
\end{align}

\paragraph{Mixed reconstructions (cross-shared + self-unique).}
\begin{align}
\hat{x}_{\mathrm{mix}} &= W_s^x z_s^y + W_u^x z_u^x + \mathbf{1}(b^x)^\top,\\
\hat{y}_{\mathrm{mix}} &= W_s^y z_s^x + W_u^y z_u^y + \mathbf{1}(b^y)^\top.
\end{align}

\subsubsection{Loss terms}
The total objective is
\begin{equation}
\mathcal{L} =
\lambda_{\mathrm{rec}}\mathcal{L}_{\mathrm{rec}}
+\lambda_{\mathrm{sh}}\mathcal{L}_{\mathrm{sh}}
+\lambda_{\mathrm{cross}}\mathcal{L}_{\mathrm{cross}}
+\lambda_{\mathrm{vic}}\mathcal{L}_{\mathrm{vic}}
+\lambda_{\mathrm{ort}}\mathcal{L}_{\mathrm{ort}}
+\lambda_{\mathrm{sp}}\mathcal{L}_{\mathrm{sp}}.
\end{equation}

\paragraph{(1) Full reconstruction.}
\begin{equation}
\mathcal{L}_{\mathrm{rec}}
= \frac{1}{Bd}\|\hat{x}-x\|_F^2+\frac{1}{Bd}\|\hat{y}-y\|_F^2 .
\end{equation}

\paragraph{(2) Shared-only reconstruction (self).}
\begin{equation}
\mathcal{L}_{\mathrm{sh}}
= \frac{1}{Bd}\|\hat{x}_s-x\|_F^2+\frac{1}{Bd}\|\hat{y}_s-y\|_F^2 .
\end{equation}

\paragraph{(3) Cross shared-only reconstruction (interchangeability).}
\begin{equation}
\mathcal{L}_{\mathrm{cross}}
= \frac{1}{Bd}\|\hat{x}_{s\leftarrow y}-x\|_F^2
 +\frac{1}{Bd}\|\hat{y}_{s\leftarrow x}-y\|_F^2 .
\end{equation}

\subsubsection{VICReg-style anti-collapse on shared latents}
We apply VICReg-style constraints~\citep{bardes2022vicreg} on $(z_s^x,z_s^y)$:
\begin{equation}
\mathcal{L}_{\mathrm{vic}}
= \alpha\,\mathcal{L}_{\mathrm{inv}}
 +\beta\left(\mathcal{L}_{\mathrm{var}}(z_s^x)+\mathcal{L}_{\mathrm{var}}(z_s^y)\right)
 +\gamma\left(\mathcal{L}_{\mathrm{cov}}(z_s^x)+\mathcal{L}_{\mathrm{cov}}(z_s^y)\right).
\end{equation}

\paragraph{Invariance.}
Let $\mathrm{BN}(\cdot)$ standardize each dimension within the minibatch (zero mean, unit std). We use
\begin{equation}
\mathcal{L}_{\mathrm{inv}}
= \frac{1}{B}\left\|\mathrm{BN}(z_s^x)-\mathrm{BN}(z_s^y)\right\|_F^2.
\end{equation}

\paragraph{Variance (hinge).}
Let $\mathrm{Std}(Z)\in\mathbb{R}^{d_s}$ denote per-dimension standard deviation across the batch. With margin $\nu>0$,
\begin{equation}
\mathcal{L}_{\mathrm{var}}(Z)
= \frac{1}{d_s}\sum_{j=1}^{d_s}\max\bigl(0,\nu-\mathrm{Std}(Z)_j\bigr)^2 .
\end{equation}

\paragraph{Covariance (decorrelation).}
Let $Z_c = Z - \frac{1}{B}\mathbf{1}\mathbf{1}^\top Z$ be batch-centered and
\begin{equation}
\mathrm{Cov}(Z) = \frac{1}{B-1} Z_c^\top Z_c .
\end{equation}
Then
\begin{equation}
\mathcal{L}_{\mathrm{cov}}(Z)
= \frac{1}{d_s}\left\|\mathrm{OffDiag}\bigl(\mathrm{Cov}(Z)\bigr)\right\|_F^2 ,
\end{equation}
where $\mathrm{OffDiag}(\cdot)$ zeros the diagonal entries.

\subsubsection{Shared--unique separability (orthogonality)}
We penalize cross-covariance between shared and unique latents within each branch. For batch-centered $A_c,B_c$,
\begin{equation}
\mathrm{Cov}(A,B) = \frac{1}{B-1}A_c^\top B_c.
\end{equation}
Then
\begin{equation}
\mathcal{L}_{\mathrm{ort}}
= \left\|\mathrm{Cov}(z_s^x,z_u^x)\right\|_F^2
 + \left\|\mathrm{Cov}(z_s^y,z_u^y)\right\|_F^2 .
\end{equation}

\subsubsection{Sparsity on unique latents}
We encourage compact residual coding via $\ell_1$ sparsity:
\begin{equation}
\mathcal{L}_{\mathrm{sp}}
= \frac{1}{B}\|z_u^x\|_{1} + \frac{1}{B}\|z_u^y\|_{1}.
\end{equation}

\subsection{SAE Metrics}
\label{app:sae_metrics}

All $R^2$ scores are computed in standardized feature space. For any reconstruction $\hat{x}$ of $x$:
\begin{align}
R^2(\hat{x};x)
&= 1 - \frac{\mathrm{MSE}(\hat{x},x)}{\mathrm{Var}(x)},\\
\mathrm{MSE}(\hat{x},x)
&= \frac{1}{Bd}\|\hat{x}-x\|_F^2,\qquad
\mathrm{Var}(x)=\frac{1}{Bd}\|x-\bar{x}\|_F^2,
\end{align}
where $\bar{x}$ is the per-dimension mean computed consistently with the standardization protocol (Appendix~\ref{app:rq1_preprocess}).

We report:
\begin{equation}
R^2_{\mathrm{full}}(x) = R^2(\hat{x};x),\qquad
R^2_{\mathrm{sh}}(x) = R^2(\hat{x}_s;x),\qquad
R^2_{\mathrm{cross}}(x) = R^2(\hat{x}_{s\leftarrow y};x),
\end{equation}
(and analogously for $y$), and define the self--cross gap
\begin{equation}
\Delta_{\mathrm{cross}}(x) = R^2_{\mathrm{sh}}(x) - R^2_{\mathrm{cross}}(x),\qquad
\Delta_{\mathrm{cross}}(y) = R^2_{\mathrm{sh}}(y) - R^2_{\mathrm{cross}}(y).
\end{equation}

\subsection{Output-Space Variance Attribution}
\label{app:sae_var}

With additive decoder contributions
\begin{equation}
x_s = W_s^x z_s^x,\qquad x_u=W_u^x z_u^x,\qquad
\varepsilon_x = x - (x_s + x_u + \mathbf{1}(b^x)^\top),
\end{equation}
we decompose (in standardized space)
\begin{equation}
\mathrm{Var}(x)
= \mathrm{Var}(x_s) + \mathrm{Var}(x_u)
+ 2\,\mathrm{Cov}(x_s,x_u) + \mathrm{Var}(\varepsilon_x),
\end{equation}
(and similarly for $y$). We report all four components; shared/unique percentages do not necessarily sum to $100\%$ when covariance/residual terms are non-zero.

\subsection{Shuffled-Pair Control for Shared-Space Saturation}
\label{app:shuffle_control}

Shared-space similarity (e.g., cosine similarity or CKA computed on the SAE shared representations) often saturates by design because the SAE objective explicitly enforces invariance between paired shared latents. To verify that this saturation is not due to a trivial solution, we perform a \emph{shuffled-pair control}: we randomly permute pairings $(x_i,y_i)$ while keeping the marginals of $x$ and $y$ unchanged, retrain SAE with the same hyperparameters, and re-compute shared-space and original-space similarity.

We expect two qualitative outcomes: (i) shared-space similarity should decrease under shuffled pairing, and (ii) original-space CKA should collapse toward zero, confirming that high shared-space alignment relies on correct pairings and is not a trivial artifact.


\subsection{SAE Hyperparameter Sweep}
\label{app:sae_sweep}

We sweep SAE settings over \texttt{use\_raw\_mse}$\in\{$False, True$\}$ and \texttt{cross\_weight}$\in\{0.0,0.1,0.2,0.5,1.0\}$, and report original-space alignment and interchangeability metrics in standardized space.

\begin{table*}[t]
\centering
\fontsize{7pt}{8pt}\selectfont
\caption{SAE sweep results (standardized space metrics). We report reconstruction quality (full/shared-only), shared-space alignment (CKA$_{\mathrm{shared}}$), and interchangeability via cross reconstruction and the self--cross gap.}
\label{tab:sae_sweep}
\resizebox{\textwidth}{!}{
\begin{tabular}{l c c c c c c c c c}
\toprule
Feature & use\_raw\_mse & cross\_weight &
$R^2_{\mathrm{full}}(x)$ & $R^2_{\mathrm{full}}(y)$ &
$R^2_{\mathrm{shared}}(x)$ & $R^2_{\mathrm{shared}}(y)$ &
CKA$_{\mathrm{shared}}$ &
$R^2_{\mathrm{cross}}(x{\leftarrow}z_s^y)$ / $R^2_{\mathrm{cross}}(y{\leftarrow}z_s^x)$ &
$\Delta_{\mathrm{cross}}(x)$ / $\Delta_{\mathrm{cross}}(y)$\\
\midrule

\multicolumn{10}{l}{\textit{Backbone features (CKA$_{\mathrm{orig}}$=0.2125)}}\\
\midrule
backbone & False & 0.0 & 0.787 & 0.901 & 0.641 & 0.784 & 0.982 & 0.492 / 0.559 & 0.149 / 0.226 \\
backbone & False & 0.1 & 0.786 & 0.902 & 0.634 & 0.784 & 0.981 & 0.537 / 0.623 & 0.098 / 0.160 \\
backbone & False & 0.2 & 0.784 & 0.900 & 0.629 & 0.768 & 0.982 & 0.541 / 0.634 & 0.087 / 0.134 \\
backbone & False & 0.5 & 0.779 & 0.897 & 0.626 & 0.768 & 0.981 & 0.582 / 0.687 & 0.045 / 0.081 \\
backbone & False & 1.0 & 0.780 & 0.896 & 0.623 & 0.772 & 0.981 & 0.598 / 0.715 & \textbf{0.025} / \textbf{0.057} \\
\cmidrule{2-10}
backbone & True  & 0.0 & 0.785 & 0.900 & 0.633 & 0.781 & 0.984 & 0.522 / 0.602 & 0.111 / 0.180 \\
backbone & True  & 0.1 & 0.783 & 0.901 & 0.623 & 0.774 & 0.983 & 0.541 / 0.627 & 0.082 / 0.146 \\
backbone & True  & 0.2 & 0.781 & 0.900 & 0.625 & 0.772 & 0.984 & 0.554 / 0.648 & 0.072 / 0.124 \\
backbone & True  & 0.5 & 0.779 & 0.892 & 0.618 & 0.763 & 0.981 & 0.579 / 0.685 & 0.039 / 0.078 \\
backbone & True  & 1.0 & 0.779 & 0.894 & 0.622 & 0.761 & 0.982 & 0.595 / 0.710 & \textbf{0.026} / \textbf{0.051} \\

\midrule
\multicolumn{10}{l}{\textit{DiT features (CKA$_{\mathrm{orig}}$=0.5369)}}\\
\midrule
dit & False & 0.0 & 0.801 & 0.926 & 0.617 & 0.825 & 0.988 & 0.534 / 0.754 & 0.071 / 0.083 \\
dit & False & 0.1 & 0.799 & 0.922 & 0.617 & 0.826 & 0.986 & 0.546 / 0.763 & 0.063 /  0.071\\
dit & False & 0.2 & 0.800 & 0.924 & 0.614 & 0.822 & 0.986 & 0.552 / 0.774 & 0.048 /  0.061\\
dit & False & 0.5 & 0.800 & 0.922 & 0.613 & 0.822 & 0.986 & 0.566 / 0.791 & 0.031 /  0.047\\
dit & False & 1.0 & 0.802 & 0.922 & 0.622 & 0.831 & 0.984 & 0.580 / 0.811 & \textbf{0.021} / \textbf{0.042} \\
\cmidrule{2-10}
dit & True  & 0.0 & 0.800 & 0.923 & 0.612 & 0.816 & 0.987 & 0.537 / 0.754 & 0.062 / 0.075 \\
dit & True  & 0.1 & 0.801 & 0.922 & 0.614 & 0.816 & 0.988 & 0.548 / 0.766 & 0.050 / 0.066 \\
dit & True  & 0.2 & 0.796 & 0.921 & 0.606 & 0.814 & 0.986 & 0.549 / 0.770 & 0.044 / 0.057 \\
dit & True  & 0.5 & 0.798 & 0.922 & 0.609 & 0.821 & 0.987 & 0.565 / 0.790 & 0.032 / 0.044 \\
dit & True  & 1.0 & 0.797 & 0.920 & 0.609 & 0.823 & 0.985 & 0.576 / 0.805 & \textbf{0.018} / \textbf{0.033} \\

\bottomrule
\end{tabular}}
\end{table*}

\subsection{Rule-based Representation-only Gating from SAE Energy Decomposition}
\label{app:gating_rules}

We build handcrafted gating rules from the Shared--Unique SAE decomposition (Appendix~\ref{app:sae_full}). For each branch, let the additive decoder contributions (in standardized feature space) be
\[
x_s = W_s^x z_s^x,\quad x_u = W_u^x z_u^x,\qquad
y_s = W_s^y z_s^y,\quad y_u = W_u^y z_u^y.
\]
We define squared-$\ell_2$ ``energy'' in shared/unique subspaces as
\[
E^{s}_{\mathrm{vlm}}=\|x_s\|_2^2,\quad E^{u}_{\mathrm{vlm}}=\|x_u\|_2^2,\qquad
E^{s}_{\mathrm{vit}}=\|y_s\|_2^2,\quad E^{u}_{\mathrm{vit}}=\|y_u\|_2^2,
\]
and $E^{\mathrm{total}}=E^s+E^u$. We use a small constant $\epsilon>0$ for numerical stability.

\paragraph{Indicators.}
We derive four indicators (symmetrically for both branches):
\[
r^{u}=\frac{E^{u}}{E^{\mathrm{total}}+\epsilon},\quad
r^{s}=\frac{E^{s}}{E^{\mathrm{total}}+\epsilon},\quad
u=\frac{E^{u}}{E^{s}+\epsilon},\quad
d=\frac{E^{s}}{E^{s}+E^{u}+\epsilon},\quad
\bar d=\frac{d_{\mathrm{vlm}}+d_{\mathrm{vit}}}{2}.
\]

\paragraph{Decision convention.}
Each strategy produces a signed score $s$ (positive favors VLM; negative favors ViT), and outputs the decision
\[
\text{choose VLM if } s > 0,\ \text{ otherwise choose ViT}.
\]

\paragraph{(i) More-unique wins.}
We compare uniqueness strength:
\[
s_1 = u_{\mathrm{vlm}} - u_{\mathrm{vit}}.
\]

\paragraph{(ii) Shared-dominant conditional (hard regime).}
Given a shared-dominance threshold $\tau$,
\[
s_2(\tau)=
\begin{cases}
r^s_{\mathrm{vlm}} - r^s_{\mathrm{vit}}, & \bar d > \tau,\\
u_{\mathrm{vlm}} - u_{\mathrm{vit}}, & \text{otherwise}.
\end{cases}
\]

\paragraph{(iii) Smoothed shared-dominance (sigmoid regime).}
To reduce sensitivity near the threshold, we replace the hard indicator by a sigmoid weight
\[
w(\bar d;\tau)=\sigma\!\bigl(\kappa(\bar d-\tau)\bigr)
=\frac{1}{1+\exp\bigl(-\kappa(\bar d-\tau)\bigr)}\in(0,1),
\]
where we fix $\kappa=5$ (corresponding to \texttt{SOFT\_LABEL\_SCALE}=5). We then define
\[
s_3(\tau)
=
w(\bar d;\tau)\,(r^s_{\mathrm{vlm}}-r^s_{\mathrm{vit}})
+
\bigl(1-w(\bar d;\tau)\bigr)\,(u_{\mathrm{vlm}}-u_{\mathrm{vit}}).
\]

\paragraph{(iv) ViT-prior fallback.}
We default to ViT and switch to VLM only when the scenario is strongly shared-dominant:
\[
\text{choose VLM if } \bar d > \tau_{\mathrm{strong}} \ \text{and}\ s_3(\tau)>0;\quad \text{else choose ViT}.
\]
This strategy is intentionally conservative to avoid over-switching. (\emph{To reproduce this variant, $\tau_{\mathrm{strong}}$ must be specified.})

\paragraph{Threshold sweep and main setting.}
We sweep the shared-dominance threshold over
\[
\tau \in \{0.5,\,0.6,\,0.7,\,0.8,\,0.9\},
\]
and select the best-performing value under the same evaluation protocol used for Table~\ref{tab:gating_main}. The main text reports results for $\tau=0.7$, which achieves the highest score among the tested thresholds. For the smoothed variant, we use the same sweep with fixed $\kappa=5$.

\subsection{Rule-based Gating: Threshold Sweep}
\label{app:gating_threshold_sweep}

We report the performance of the shared-dominance threshold sweep for the rule-based gates. The main text uses $\tau=0.7$.

\begin{table}[t]
\centering
\small
\caption{Threshold sweep for rule-based gating. Fill with the same evaluation metrics used in Table~\ref{tab:gating_main}.}
\label{tab:gating_tau_sweep}
\begin{tabular}{c c c c c}
\toprule
$\tau$ & navtest PDMS(\%) \\
\midrule
0.5 & 89.73  \\
0.6 & 89.75  \\
0.7 & 89.92  \\
0.8 & 89.90  \\
0.9 & 89.89  \\
\bottomrule
\end{tabular}
\end{table}

\subsection{Learned Representation-only Gating Models and t-SNE Diagnostics}
\label{app:gating_models}

We formulate gating as supervised learning. For each scenario, we construct inputs from both branches' representations at either the backbone level or the DiT level. Common feature constructions include concatenation $[x;y]$, difference $(x-y)$, and their combination $[x;y;x-y]$.

\paragraph{Labels.}
The binary label indicates which branch yields better closed-loop performance for that scenario (VLM-better vs.\ ViT-better), computed from the evaluation score used in Table~\ref{tab:gating_main}. Ties can be discarded or broken deterministically.

\paragraph{Model families.}
We evaluate:
\begin{itemize}
    \item \textbf{Tree-based models:} Random Forest; Gradient Boosting / GBDT variants.
    \item \textbf{MLP gate:} multi-layer fully-connected network with BatchNorm and Dropout, trained with binary cross-entropy, sigmoid output.
    \item \textbf{Token-aware attention gate:} self-attention encoder over the VLM token sequence; cross-attention using the vision-only global feature as a query over VLM keys/values; MLP head for classification. This explicitly uses the long VLM token sequence rather than pooled features alone.
\end{itemize}

\begin{figure}[t]
    \centering
    \includegraphics[width=1\linewidth]{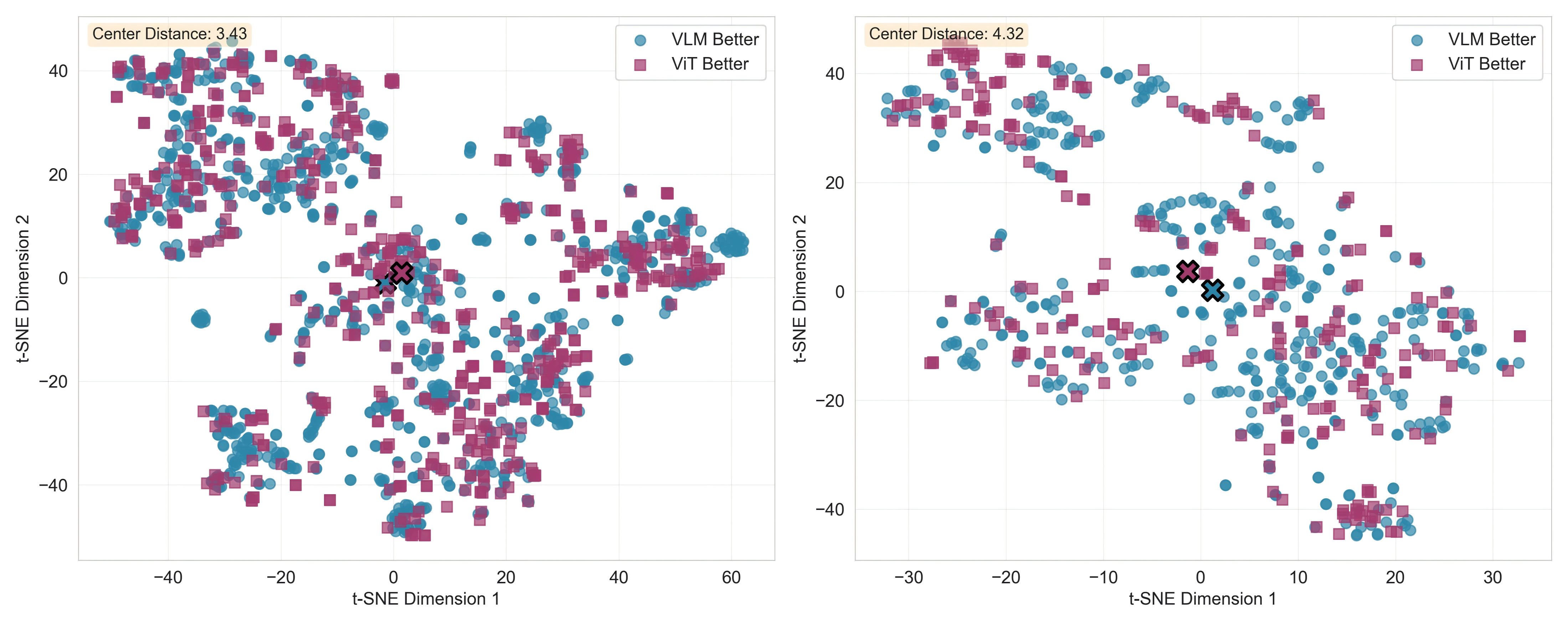}
    \caption{t-SNE of backbone- and DiT-level features colored by whether VLM outperforms ViT for the scenario. The classes are not separable, suggesting intrinsic difficulty for representation-only gating. (Left: backbone-level features; Right: DiT-level features.)}
    \label{fig:tsne}
\end{figure}

\paragraph{t-SNE diagnostic.}
To qualitatively assess separability, we apply t-SNE on backbone-level and DiT-level representations and color points by the binary label (VLM-better vs.\ ViT-better). Poor class separation in both spaces (Fig.~\ref{fig:tsne}) is consistent with the difficulty of representation-only gating, though t-SNE is used only as a visualization tool rather than a definitive test.

\section{Offline Trajectory-quality Score in NAVSIM (PDMS v1 and EPDMS v2)}
\label{app:offline_score}

We use the planning-oriented NAVSIM benchmark and adopt the official Predictive Driver Model Score (PDMS) from NAVSIM v1 as our primary offline trajectory-quality score $s(\cdot,\cdot)$; higher is better. PDMS is a pseudo closed-loop metric that holistically assesses safety, comfort, and progress via multiplicative penalties and a weighted average:
\begin{equation}
\label{eq:pdms}
\mathrm{PDMS} \;=\; \mathrm{NC} \times \mathrm{DAC} \times
\left(\frac{5\cdot \mathrm{EP} + 5\cdot \mathrm{TTC} + 2\cdot \mathrm{C}}{12}\right),
\end{equation}
where $\mathrm{NC}$ denotes no at-fault collisions, $\mathrm{DAC}$ drivable-area compliance, $\mathrm{EP}$ ego progress, $\mathrm{TTC}$ time-to-collision within bound, and $\mathrm{C}$ comfort.

\paragraph{NAVSIM v2: Extended PDMS (EPDMS).}
NAVSIM v2 extends PDMS to improve coverage and fairness of open-loop planning evaluation. Compared to NAVSIM v1, EPDMS introduces additional weighted subscores (lane keeping and extended comfort variants), additional multiplier penalties (driving direction compliance and traffic light compliance), and a false-positive penalty filtering scheme.

\begin{table}[t]
\centering
\small
\caption{EPDMS composition in NAVSIM v2 (new metrics relative to v1 are highlighted).}
\label{tab:epdms_metrics}
\begin{tabular}{l c c}
\toprule
Metric & Weight & Range \\
\midrule
No at-fault Collisions (NC) & multiplier & $\{0, \tfrac{1}{2}, 1\}$ \\
Drivable Area Compliance (DAC) & multiplier & $\{0, 1\}$ \\
\textbf{Driving Direction Compliance (DDC)} & multiplier & $\{0, \tfrac{1}{2}, 1\}$ \\
\textbf{Traffic Light Compliance (TLC)} & multiplier & $\{0, 1\}$ \\
Ego Progress (EP) & 5 & $[0,1]$ \\
Time to Collision (TTC) within bound & 5 & $\{0,1\}$ \\
\textbf{Lane Keeping (LK)} & 2 & $\{0,1\}$ \\
\textbf{History Comfort (HC)} & 2 & $\{0,1\}$ \\
\textbf{Extended Comfort (EC)} & 2 & $\{0,1\}$ \\
\bottomrule
\end{tabular}
\end{table}

\paragraph{False-positive penalty filtering.}
To reduce false-positive penalties, NAVSIM v2 disables a penalty when the human agent is also responsible for the corresponding violation. Formally, for a metric $m$, define
\begin{equation}
\label{eq:epdms_filter}
\mathrm{filter}_m(\mathrm{agent},\mathrm{human}) =
\begin{cases}
1.0, & \text{if } m(\mathrm{human}) = 0,\\
m(\mathrm{agent}), & \text{otherwise.}
\end{cases}
\end{equation}
Intuitively, if the human baseline also triggers the violation, the metric is neutralized (set to $1.0$) rather than penalizing the planner.

\paragraph{EPDMS definition.}
With the above filtering, EPDMS is defined as
\begin{align}
\label{eq:epdms}
\mathrm{EPDMS} \;=\;&
\left(\prod_{m\in\{\mathrm{NC},\mathrm{DAC},\mathrm{DDC},\mathrm{TLC}\}} \mathrm{filter}_m(\mathrm{agent},\mathrm{human})\right)
\\
&\cdot
\left(
\frac{\sum_{m\in\{\mathrm{TTC},\mathrm{EP},\mathrm{HC},\mathrm{LK},\mathrm{EC}\}} w_m \cdot \mathrm{filter}_m(\mathrm{agent},\mathrm{human})}
{\sum_{m\in\{\mathrm{TTC},\mathrm{EP},\mathrm{HC},\mathrm{LK},\mathrm{EC}\}} w_m}
\right),
\end{align}
where $w_m$ are the weights listed in Table~\ref{tab:epdms_metrics}.

\paragraph{Pseudo closed-loop aggregation in NAVSIM v2.}
NAVSIM v1 computes metrics after a 4-second non-reactive simulation rollout (background actors follow recorded futures; ego follows the planned trajectory via a controller). NAVSIM v2 uses a two-stage aggregation to better approximate closed-loop behavior while remaining open-loop:
(i) a first-stage score is computed on an initial 4-second scene;
(ii) multiple follow-up scenes (precomputed rollouts starting from the same initial scene but with different end states) are also scored, and then aggregated with weights given by a Gaussian kernel based on how close each follow-up scene's start state is to the submitted planner's first-stage end state.
Finally, the first-stage score and the aggregated second-stage score are multiplied to obtain the final aggregated EPDMS score.

\section{RQ2 Win Counting Protocol and Thresholds}
\label{app:rq2_win_count}

For scenario-level complementarity, we compare policies via the per-scenario advantage
\[
\Delta_{r,i} = s(\mathrm{VLM},r,i)-s(\mathrm{ViT},r,i),
\]
where $s(m,r,i)$ is the NAVSIM v1 PDMS score of policy $m$ on scenario $i$ under random seed $r\in\{1,2,3\}$ (Appendix~\ref{app:offline_score}). A per-seed significant win is defined as $|\Delta_{r,i}|>\tau$.

\section{Additional Cross-model Evidence Beyond InternVL3-2B}
\label{app:extra_generalization}

To test whether the shared-plus-unique structure is specific to a single VLM, we additionally trained InternVL3-8B and Qwen3VL-8B on NAVSIM with the same three-stage recipe as the main model and repeated the RQ1 analysis. We also repeated the same style of analysis on AsyncDrive/nuPlan, using GameFormer as the fast branch and Llama2-13B as the slow branch. For the NAVSIM runs, the three CKA values are reported in the order of VLM paired with ViT, ResNet, and EVA-CLIP. Specifically, for InternVL3-8B, backbone CKA is $0.26/0.23/0.24$ and DiT CKA is $0.49/0.48/0.48$; for Qwen3VL-8B, backbone CKA is $0.30/0.29/0.31$ and DiT CKA is $0.55/0.54/0.54$. On AsyncDrive/nuPlan, the rebuttal analysis reported a single planner-level CKA summary of $0.66$.

\begin{table*}[t]
\centering
\fontsize{8pt}{9.5pt}\selectfont
\setlength{\tabcolsep}{4pt}
\renewcommand{\arraystretch}{0.96}
\caption{Additional SAE/CCA evidence for RQ1 across larger VLMs and an external planner stack. Higher $R^2_{\mathrm{cross}}$ and CCA indicate stronger shared structure. For AsyncDrive/nuPlan, the rebuttal analysis reported only a single planner-level summary rather than separate backbone and DiT values.}
\label{tab:extra_generalization}
\begin{adjustbox}{max width=\linewidth}
\begin{tabular}{@{}lccccc@{}}
\toprule
Setting & PDMS & $R^2_{\mathrm{cross}}$ (backbone) & $R^2_{\mathrm{cross}}$ (DiT / planner) & CCA mean@10 & Note \\
\midrule
InternVL3-8B on NAVSIM & 90.4 & 0.49 / 0.62 & 0.54 / 0.78 & 0.81 / 0.97 & CKA in text \\
Qwen3VL-8B on NAVSIM & 90.7 & 0.44 / 0.60 & 0.52 / 0.81 & 0.84 / 0.95 & CKA in text \\
AsyncDrive / nuPlan & - & - & 0.61 / 0.73 & - / 0.88 & planner-level summary only \\
\bottomrule
\end{tabular}
\end{adjustbox}
\end{table*}

\section{Token-level robustness to mean pooling.}
\label{app:Token-level}

Because the main paper uses pooled VLM features for tractable full-dataset statistics, we additionally ran token-level / non-global-pooling analyses over 20 random trials on the largest subset that fits our hardware (roughly one-tenth of the full dataset). CKA and SAE statistics remain within about $5\%$ of the main values, while CCA mean@10 fluctuates by roughly $5$--$10\%$ in some runs. Crucially, the qualitative conclusion is unchanged: VLM and vision-only backbones differ more strongly at the backbone level, become more aligned after policy learning, and still preserve non-shared residual factors.

\section{Domain-Adaptation Control}
\label{app:domain_adaptation_control}

\paragraph{Motivation.}
The main experiments compare practical VLA-style and vision-only E2E representation pipelines under the same downstream planner.
However, the upstream histories are not identical: the VLM branch uses a driving-domain adaptation stage, whereas the vision-only branch starts from a strong public visual checkpoint and is trained through the downstream planner.
We therefore add a data-restricted control to test whether the observed VLM--ViT complementarity is mainly caused by this domain-adaptation asymmetry.

\paragraph{Controlled setups.}
We evaluate two additional variants.
First, \textbf{VLM-NAVSIM-DA} restricts the VLM driving adaptation stage to NAVSIM training data and its available text/trajectory annotations, instead of using the full multi-dataset RecogDrive-style corpus.
The adapted VLM encoder is then frozen and inserted into the same downstream planner training pipeline as the original VLM branch.
Second, \textbf{ViT-NAVSIM-DA} initializes from the same public ViT-large checkpoint as the original vision-only baseline, performs MAE adaptation on NAVSIM training images only, discards the MAE decoder, and then follows the same downstream planner training protocol as the original ViT-large branch.
All variants use the same NAVSIM split, trajectory decoding format, downstream planner/action architecture, and evaluation protocol.
No NAVSIM test images or evaluation labels are used in the adaptation stage.

\paragraph{Why ViT is not trained from scratch.}
We initialize ViT-NAVSIM-DA from the public ViT-large checkpoint rather than training it from scratch.
The purpose of this control is to test the effect of \emph{in-domain adaptation}, not to remove generic large-scale visual pretraining.
Training ViT from scratch on NAVSIM would introduce a much stronger and less realistic pretraining-budget asymmetry, since the VLM branch necessarily starts from a large pretrained multimodal checkpoint.

\paragraph{Results.}
Table~\ref{tab:domain_adaptation_control} summarizes the control.
Restricting the VLM adaptation to NAVSIM causes only a small performance change compared with VLM-full-DA.
MAE adaptation also gives the vision-only branch a comparable in-domain adaptation stage.
More importantly, the VLM--ViT complementarity remains: the VLM branch still has a clear advantage in complex semantic and interaction-heavy scenes, while the cross-model best-of-two score remains substantially above either single branch.
This suggests that the observed complementarity is not explained away by additional driving-domain exposure alone.

\begin{table}[t]
\centering
\caption{
\textbf{Domain-adaptation control on NAVSIM.}
\textit{Main} denotes VLM-full-DA vs.\ ViT-large.
\textit{NAVSIM-DA} denotes VLM-NAVSIM-DA vs.\ ViT-NAVSIM-DA.
}
\label{tab:domain_adaptation_control}
\fontsize{8pt}{9.5pt}\selectfont
\setlength{\tabcolsep}{4pt}
\renewcommand{\arraystretch}{0.95}
\begin{tabular}{@{}lcccc@{}}
\toprule
Setting
& VLM PDMS $\uparrow$
& ViT PDMS $\uparrow$
& Oracle $\uparrow$
& Complex wins \\
\midrule
Main
& 90.80
& 88.88
& 93.58
& 139 / 187 \\
NAVSIM-DA
& 90.64
& 87.45
& 93.20
& 133 / 187 \\
\bottomrule
\end{tabular}
\end{table}

\paragraph{Interpretation and scope.}
This control should not be interpreted as a fully symmetric pretraining comparison.
VLM and ViT differ in architecture, modality, and upstream objectives, and a perfectly matched adaptation budget is not well-defined.
Instead, the control addresses a concrete confound: whether the complementarity observed in the main experiments disappears once the vision-only branch also receives in-domain adaptation and the VLM adaptation is restricted to the same dataset.
The persistence of the complex-scene advantage and the cross-model oracle gain indicates that the main conclusion is not an artifact of extra driving-domain data alone.

\section{Scene Taxonomy for Major-Decision-Mismatch Cases}
\label{app:rq2_taxonomy}

\paragraph{Motivation.}
RQ2 analyzes 279 major-decision-mismatch cases, where one branch makes a substantially worse decision while the other remains broadly reasonable under a strict threshold $\tau=0.9$.
Because these cases are heterogeneous, we use a coarse scene taxonomy to understand where VLM--vision complementarity arises.
The taxonomy is intended for aggregate behavioral analysis rather than fine-grained semantic benchmarking.

\paragraph{Scene-description generation.}
For each of the 279 cases, we first generate a driving-scene description following the VLM annotation format used in RecogDrive-style NAVSIM training data.
Given the front-view image, ego history, navigation command, and available scene metadata, the annotation prompts ask for planning-relevant factors such as surrounding vehicles and pedestrians, road markings, traffic lights, congestion, vulnerable road users, influential objects, and the likely short-horizon driving behavior.
An example annotation includes scene-level descriptions, congestion status, road-marking interpretation, influential-object analysis, vulnerable-road-user detection, and a short speed/path decision explanation.
These descriptions are used only for post-hoc scenario grouping and are not used to train the policies or the trajectory scorer.

\paragraph{Coarse grouping protocol.}
We then use GPT-5.4 to assign each case to one primary scene type based on the generated description and the driving context.
When multiple factors are present, we assign the type that is most relevant to the trajectory decision difference between the two branches.
The authors manually checked the AI-assisted annotations and found no obvious errors or omissions.
The six coarse categories are:

\begin{table}[h]
\centering
\caption{
\textbf{Coarse scene taxonomy for the 279 major-decision-mismatch cases.}
The taxonomy is used only for RQ2 behavioral analysis.
}
\label{tab:rq2_taxonomy_definition}
\small
\setlength{\tabcolsep}{4pt}
\renewcommand{\arraystretch}{0.95}
\begin{tabular}{@{}lp{0.66\linewidth}@{}}
\toprule
Category & Definition \\
\midrule
Simple lane keeping &
Straightforward lane-following cases with no salient semantic ambiguity, unusual object, or interaction-heavy event. \\
Intersection semantics &
Cases where traffic lights, intersections, crossing structure, right-of-way, or turn/straight semantics strongly influence the decision. \\
Curve / merge / ramp &
Cases involving curved roads, ramp-like geometry, lane merging, or route-following ambiguity. \\
Work zone / cones &
Cases where cones, construction areas, blocked lanes, or temporary traffic layouts affect the feasible path. \\
Narrow / unclear boundary &
Cases with narrow drivable space, weak or ambiguous lane boundaries, parked vehicles near the ego path, or unclear road edges. \\
Occlusion / dense clutter &
Cases with occluded agents, dense traffic, pedestrians, cluttered urban layouts, or limited visibility of relevant interactions. \\
\bottomrule
\end{tabular}
\end{table}

\paragraph{Simple vs.\ complex split.}
We additionally merge the six categories into a binary simple/complex split.
A case is labeled \emph{simple} only when the required trajectory decision is straight lane following and the scene description contains no obvious out-of-distribution object, temporary event, semantic ambiguity, dense interaction, or unclear drivable boundary.
All other cases are labeled \emph{complex}.
This definition is intentionally conservative: rather than trying to precisely quantify all forms of complexity, we only mark a scene as simple when it is clearly simple.
Under this protocol, the simple subset contains 92 cases and the complex subset contains 187 cases.

\paragraph{Use in RQ2.}
This taxonomy yields the scene-level breakdown reported in Sec.~\ref{sec:rq2_scenario}.
In the 279 major-decision-mismatch cases, VLM wins 176 cases and ViT wins 103.
ViT wins more often only in simple lane keeping, whereas VLM dominates the semantically complex and interaction-heavy categories.
The taxonomy should therefore be interpreted as a coarse diagnostic of where the two branches differ, not as a claim that the six categories exhaust all possible driving factors.

\section{Scenario Taxonomy for Major-difference Cases}
\label{app:scenario_taxonomy}

For the semantic scenario-breakdown analysis in RQ2, we use a stricter threshold $\tau=0.9$ than the stability-oriented win-counting thresholds in the main text. This subset is intended to isolate cases where one branch makes a major decision error while the other remains broadly reasonable. Under this criterion, we obtain 279 scenarios, of which VLM wins 176 and ViT wins 103.

\begin{table*}[t]
\centering
\fontsize{8pt}{9.5pt}\selectfont
\setlength{\tabcolsep}{4pt}
\renewcommand{\arraystretch}{0.96}
\caption{Scenario taxonomy for the 279 major-difference cases ($\tau=0.9$). Percentages are relative to the 279 categorized cases.}
\label{tab:scenario_taxonomy}
\begin{tabular}{@{}lcccccc@{}}
\toprule
Scenario type & Count & Share & VLM better & ViT better & VLM win rate & Complexity \\
\midrule
simple\_lane\_keep & 92 & 33.0\% & 37 & 55 & 40.2\% & Simple \\
intersection\_semantic & 78 & 28.0\% & 58 & 20 & 74.4\% & Complex \\
curve\_merge\_ramp & 38 & 13.6\% & 23 & 15 & 60.5\% & Complex \\
workzone\_cone & 15 & \phantom{0}5.4\% & 12 & \phantom{0}3 & 80.0\% & Complex \\
narrow\_unclear\_boundary & 44 & 15.8\% & 34 & 10 & 77.3\% & Complex \\
occlusion\_dense\_clutter & 12 & \phantom{0}4.3\% & 12 & \phantom{0}0 & 100.0\% & Complex \\
\midrule
Simple (merged) & 92 & 33.0\% & 37 & 55 & 40.2\% & - \\
Complex (merged) & 187 & 67.0\% & 139 & 48 & 74.3\% & - \\
\bottomrule
\end{tabular}
\end{table*}

The long-tail categories most favorable to the VLM branch are intersections / semantic decisions ($58/78$), work zones / cones ($12/15$), unclear boundaries / narrow roads ($34/44$), and occlusion / dense clutter ($12/12$). By contrast, ViT wins more often in the simple lane-keeping subset ($55/92$), consistent with the main-text claim that VLM-specific residual factors are particularly useful in scenarios that require semantic interpretation, ambiguity resolution, or stronger multi-agent interaction handling.

\section{Behavioral Statistics and Safety Diagnostics}
\label{app:behavior_safety}

Table~\ref{tab:behavior_safety} provides the full behavioral and safety diagnostics summarized in RQ2, with standard deviations computed over five random seeds.
The VLM branch is not only faster on average, but also exhibits a distinct progress--braking trade-off: it advances further while showing stronger braking-side responses when required.

\begin{table*}[t]
\centering
\fontsize{8pt}{9.5pt}\selectfont
\setlength{\tabcolsep}{4pt}
\renewcommand{\arraystretch}{0.96}
\caption{Behavioral and safety diagnostics for VLM vs.\ ViT on NAVSIM. The significant-difference subset contains scenarios where the two branches differ substantially in score.}
\label{tab:behavior_safety}
\begin{adjustbox}{max width=\linewidth}
\begin{tabular}{@{}lcccc@{}}
\toprule
Metric & VLM (full set) & ViT (full set) & VLM (significant subset) & ViT (significant subset) \\
\midrule
Mean path speed (m/s) & 5.31($\pm$ 0.05) & 5.17($\pm$ 0.02) & 5.74($\pm$ 0.11) & 5.43($\pm$ 0.04) \\
Path length (m) & 18.60($\pm$ 0.07) & 18.09($\pm$ 0.05) & 20.08($\pm$ 0.14) & 19.02($\pm$ 0.11) \\
Min longitudinal acceleration (m/s$^2$) & -2.87($\pm$ 0.01) & -2.69($\pm$ 0.01) & -2.99($\pm$ 0.03) & -2.53($\pm$ 0.02) \\
Cumulative deceleration & 4.94($\pm$ 0.02) & 4.63($\pm$ 0.02) & 5.26($\pm$ 0.02) & 4.38($\pm$ 0.02) \\
Hard-brake segments & 0.49($\pm$ 0.01) & 0.44($\pm$ 0.00) & 0.49($\pm$ 0.01) & 0.36($\pm$ 0.00) \\
No-collision score (NC) & 97.9($\pm$ 0.01) & 97.5($\pm$ 0.01) & - & - \\
Drivable-area compliance (DAC) & 97.3($\pm$ 0.01) & 97.1($\pm$ 0.01) & - & - \\
Time-to-collision (TTC) & 94.9($\pm$ 0.01) & 93.6($\pm$ 0.01) & - & - \\
Minimum obstacle distance (m) & 2.12($\pm$ 0.02) & 2.11($\pm$ 0.02) & 2.7($\pm$ 0.02) & 1.8($\pm$ 0.01) \\
\bottomrule
\end{tabular}
\end{adjustbox}
\end{table*}

\paragraph{Intervention-style evidence from SAE-decomposed latents.}
We also tested whether the residual factors isolated by the Shared--Unique SAE are behaviorally useful by training policies on different latent combinations. On NAVSIM, shared-only features reach $83.0$ PDMS; shared + ViT-unique reaches $86.1$; shared + VLM-unique reaches $86.9$; and shared + both unique reaches $87.5$. The effect is even clearer in the 187 semantically complex / interaction-heavy cases: shared + VLM-unique yields high-score outcomes in $86$ scenes, compared with $37$ for shared + ViT-unique and $33$ for shared-only. This supports the interpretation that VLM-specific residual factors are especially useful in complex scenarios rather than being purely representational artifacts.

\paragraph{Tail-risk of the scorer-based switcher.}
We do not interpret the scorer as a certified safety module. Instead, we use it as a practical selector whose main operating knob is the fallback threshold: raising the threshold invokes the slow VLM branch more often, improving worst-case behavior at the cost of latency. To quantify the residual risk, we count \emph{severe failures}: cases where the ViT and VLM trajectories have a large score gap but the system still fails to fall back to the VLM branch.

\begin{table}[t]
\centering
\fontsize{8pt}{9.5pt}\selectfont
\setlength{\tabcolsep}{4pt}
\renewcommand{\arraystretch}{0.96}
\caption{Tail-risk analysis of severe switcher failures. Exact VLM invocation for the more conservative setting was not reported in the rebuttal, so it is marked with ``-''.}
\label{tab:scorer_tail_risk}
\begin{tabular}{@{}lccc@{}}
\toprule
Setting & Severe failures ($>20\%$ gap) & Severe failures ($>50\%$ gap) & VLM invocation \\
\midrule
DualDriveVLA (default) & \phantom{0}50 & 30 & about 15\% \\
More conservative fallback & \phantom{0}40 & 21 & - \\
\bottomrule
\end{tabular}
\end{table}

\section{Oracle Controls Across Model Families}
\label{app:oracle_controls}

To test whether the gains in RQ2/RQ3 are simply due to generic ensemble diversity, we also computed oracle controls across vision--vision and VLM--VLM pairings. Their gains are substantially smaller than those of VLM--vision pairing, suggesting that not all diversity is equally useful.

\begin{center}
\fontsize{8pt}{9.5pt}\selectfont
\setlength{\tabcolsep}{4pt}
\renewcommand{\arraystretch}{0.96}
\captionof{table}{Oracle controls across model families. Gains are measured relative to the best single model inside each candidate set.}
\label{tab:oracle_controls}
\begin{tabular}{@{}lcc@{}}
\toprule
Candidate set & Oracle score & Gain \\
\midrule
ViT + ResNet & 89.75 & +0.87 \\
ViT + EVA-CLIP & 89.61 & +0.73 \\
ResNet + EVA-CLIP & 88.69 & +0.80 \\
ViT + ResNet + EVA-CLIP & 90.05 & +1.17 \\
InternVL + Qwen & 91.90 & +1.10 \\
VLM + ViT & 93.58 & +2.78 \\
\bottomrule
\end{tabular}
\end{center}

\section{Qualitative Case Gallery (VLM vs.\ ViT vs.\ Expert)}
\label{app:rq2_cases}

\begin{figure}[h!]
  \centering
  \begin{minipage}[t]{0.48\textwidth}
    \centering
    \includegraphics[width=\linewidth]{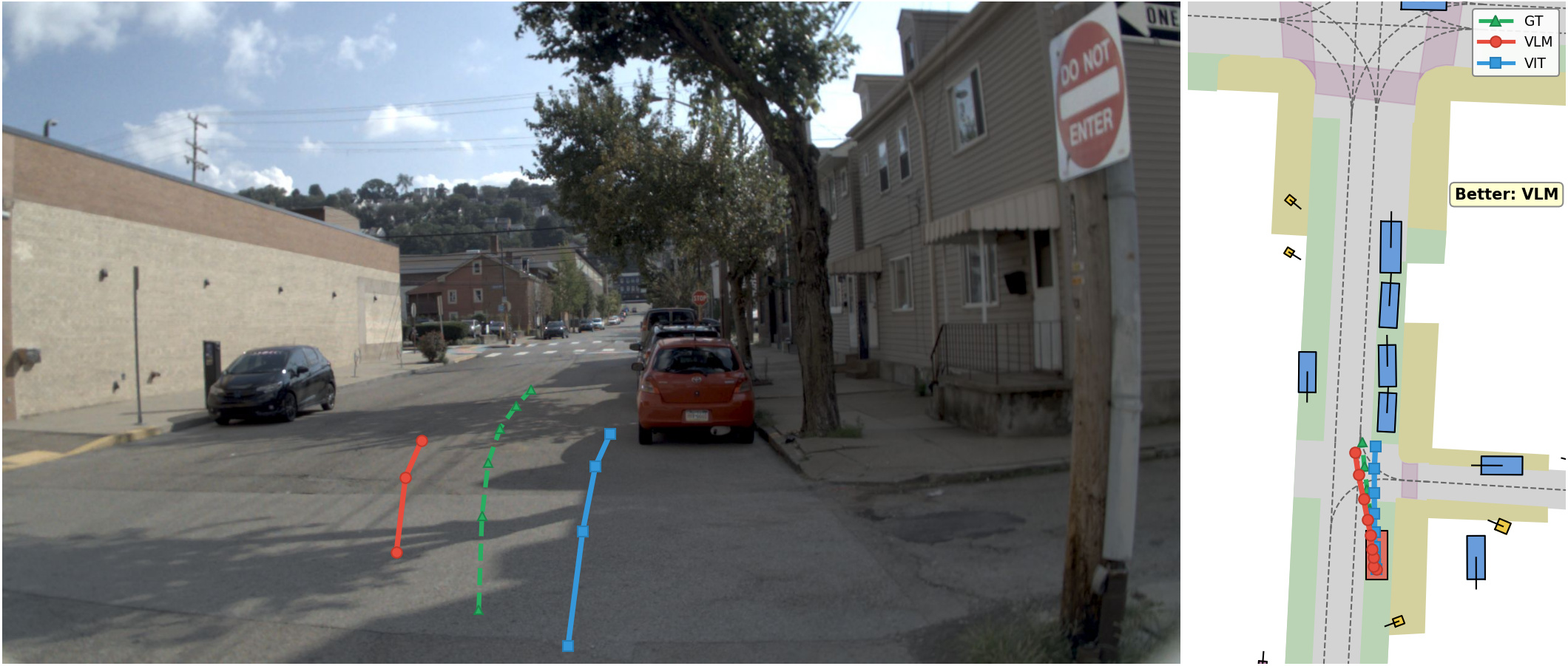}\\[-1mm]
    \includegraphics[width=\linewidth]{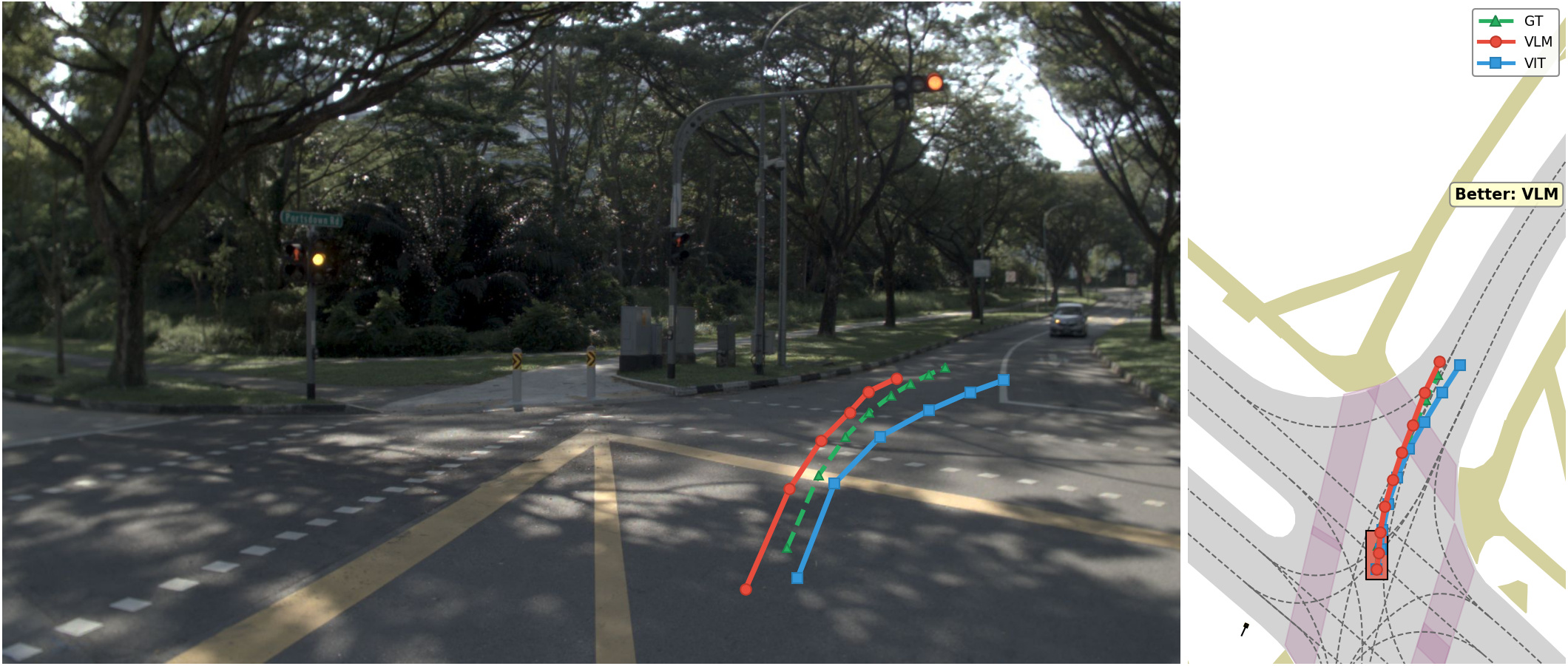}\\[-1mm]
    \includegraphics[width=\linewidth]{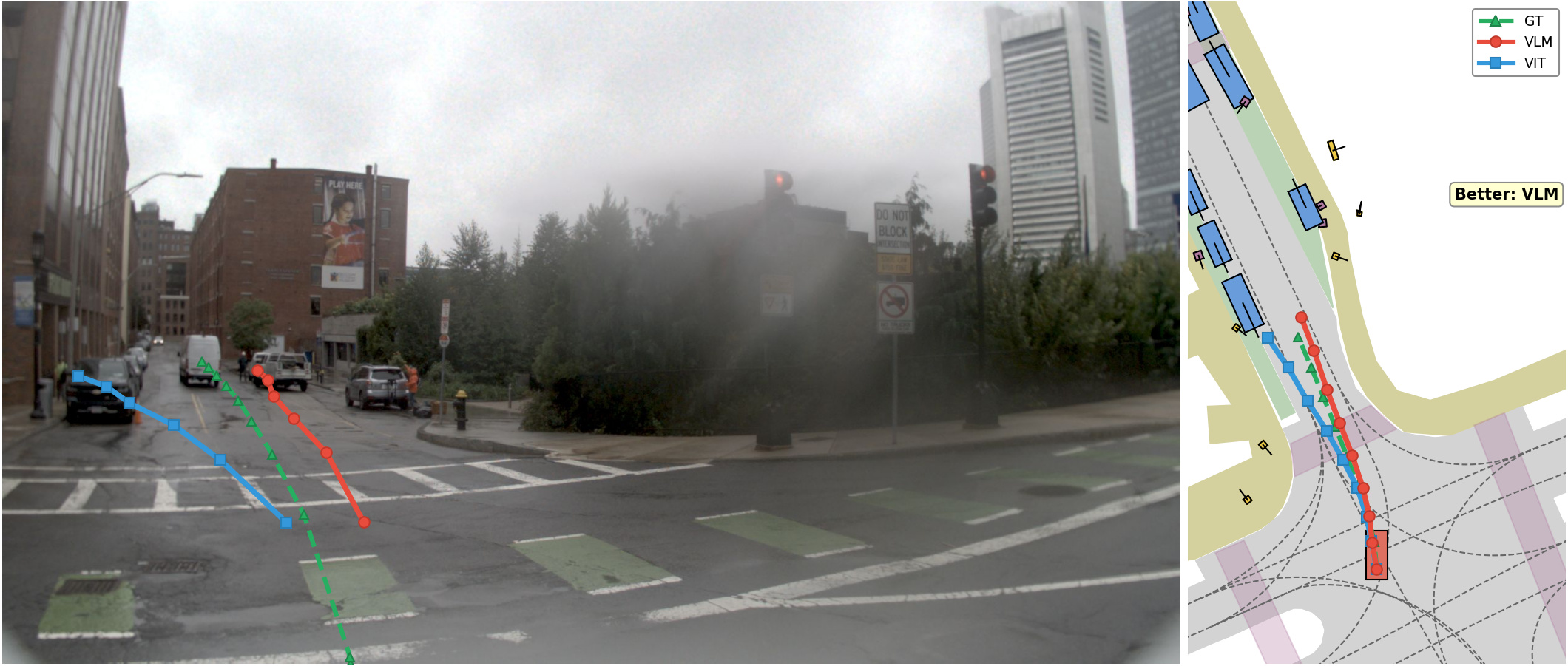}\\[-1mm]
    \includegraphics[width=\linewidth]{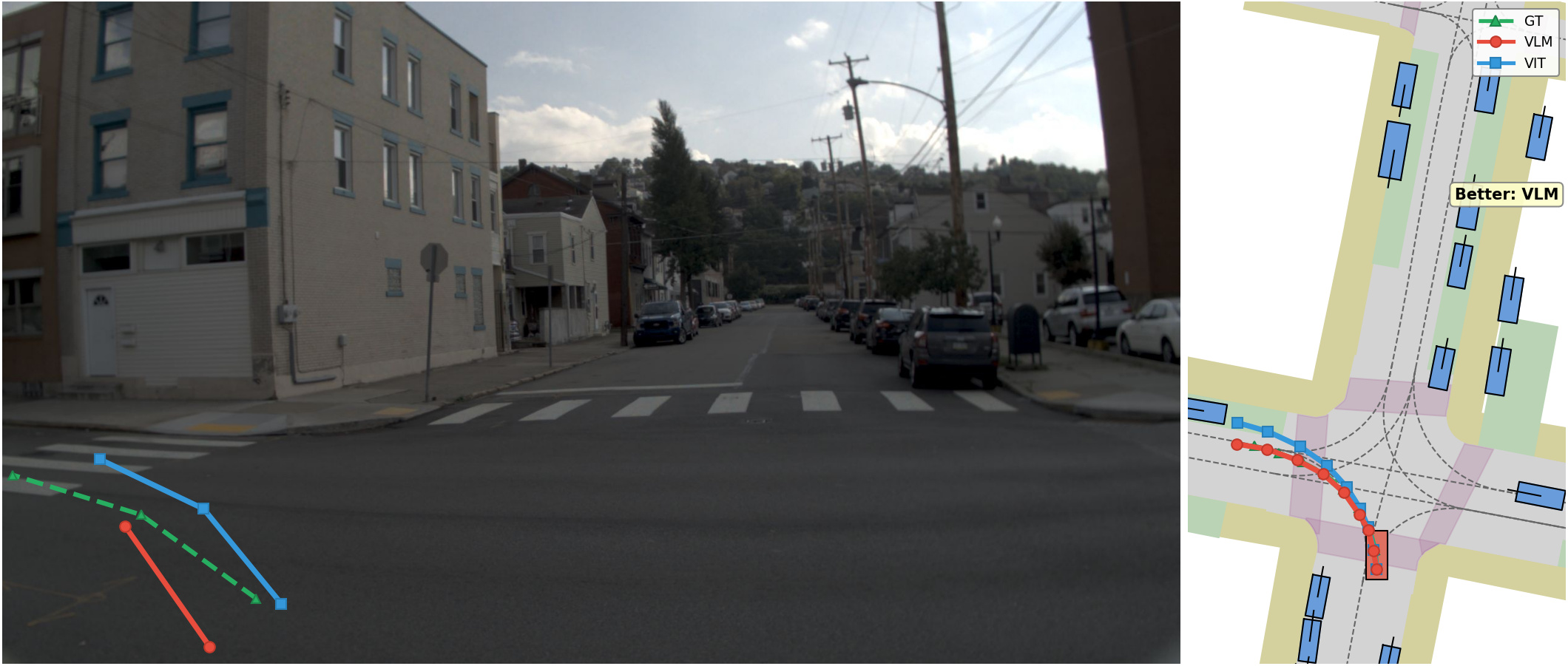}\\[-1mm]
    \includegraphics[width=\linewidth]{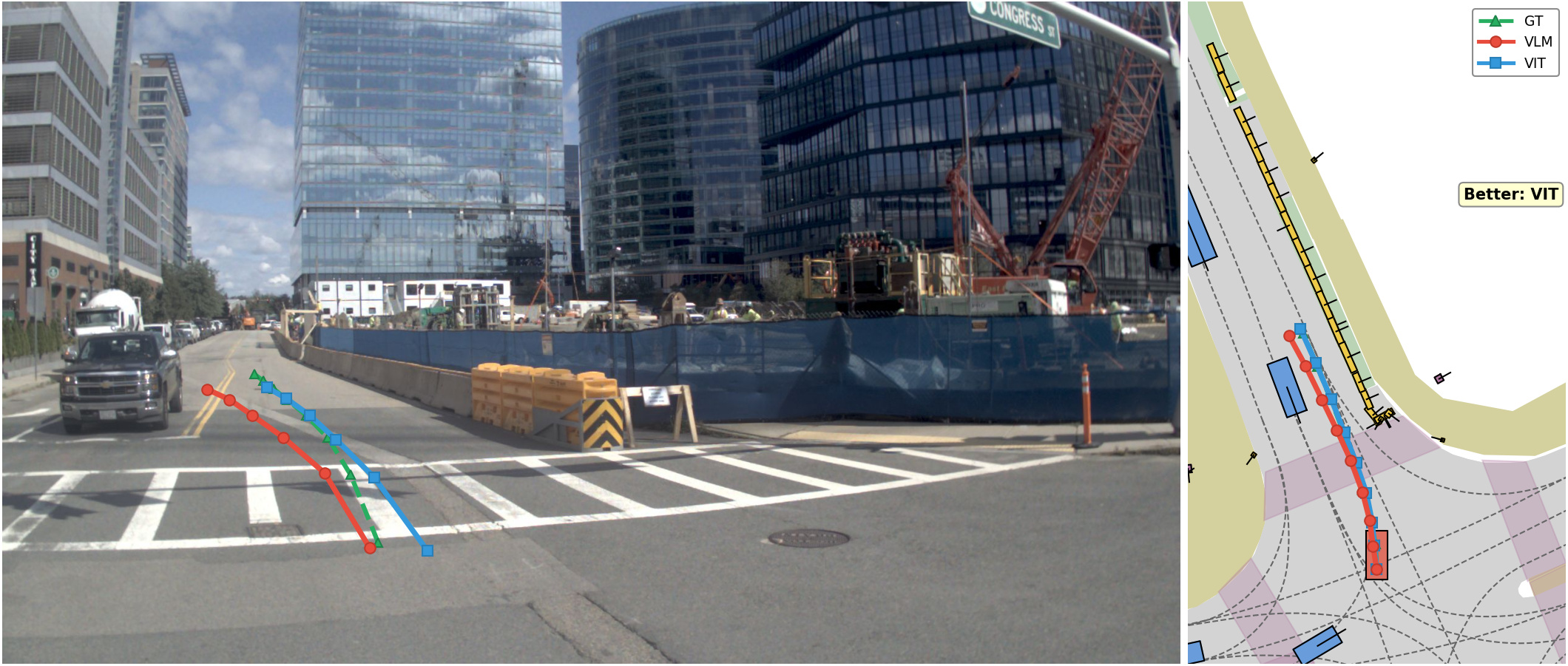}
  \end{minipage}\hfill
  \begin{minipage}[t]{0.48\textwidth}
    \centering
    \includegraphics[width=\linewidth]{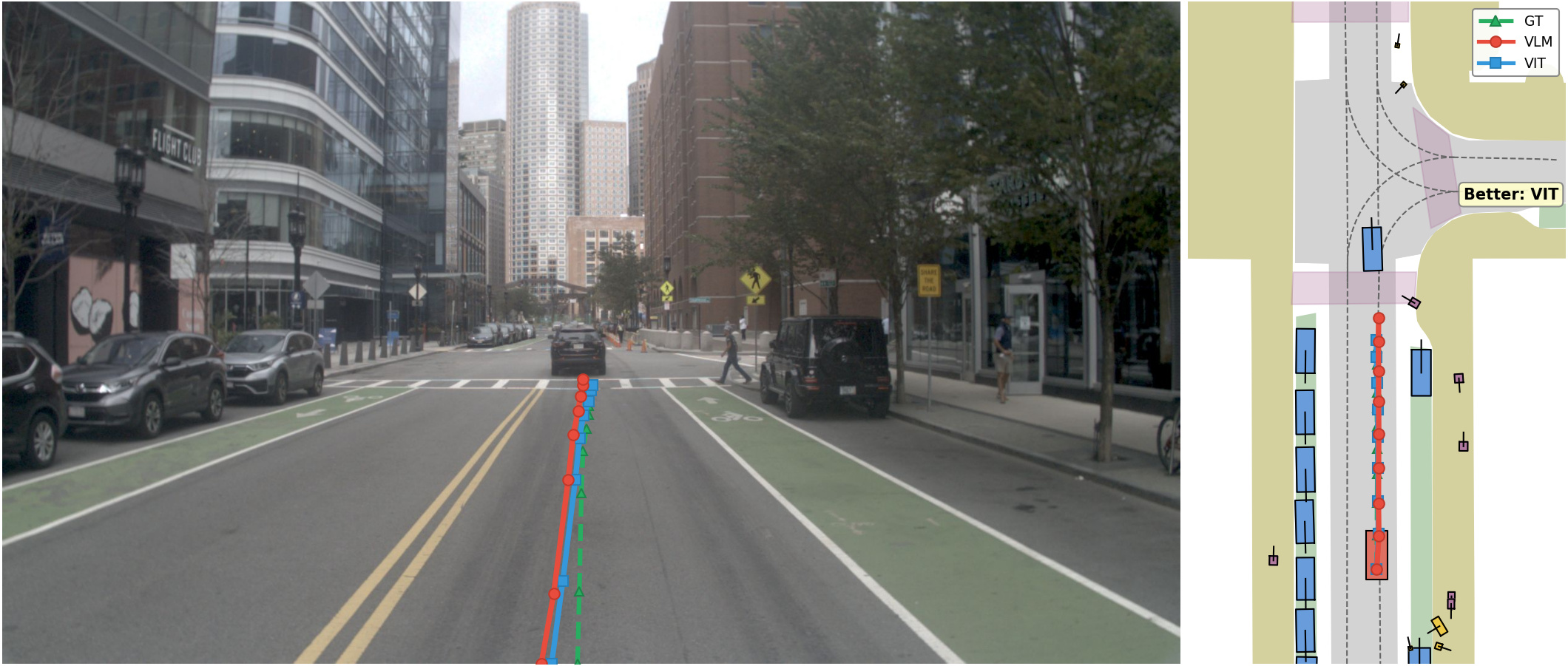}\\[-1mm]
    \includegraphics[width=\linewidth]{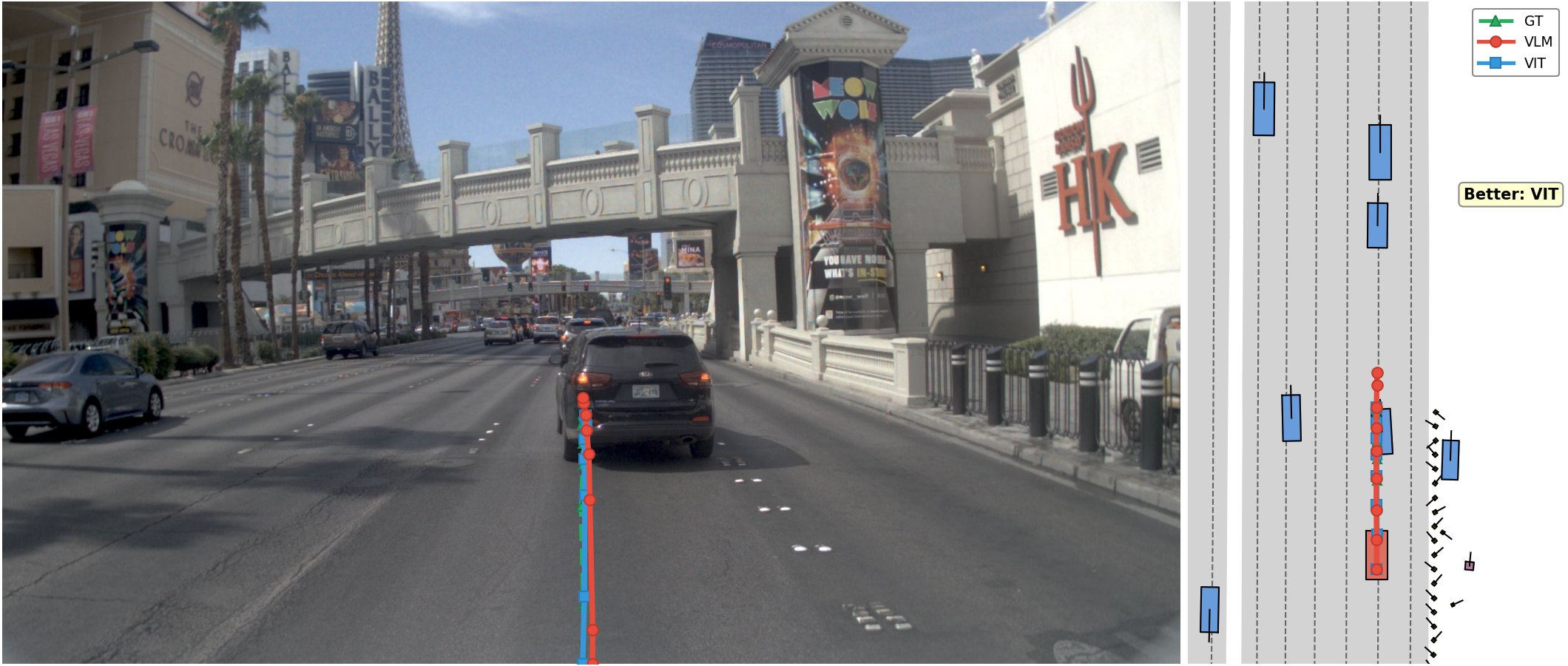}\\[-1mm]
    \includegraphics[width=\linewidth]{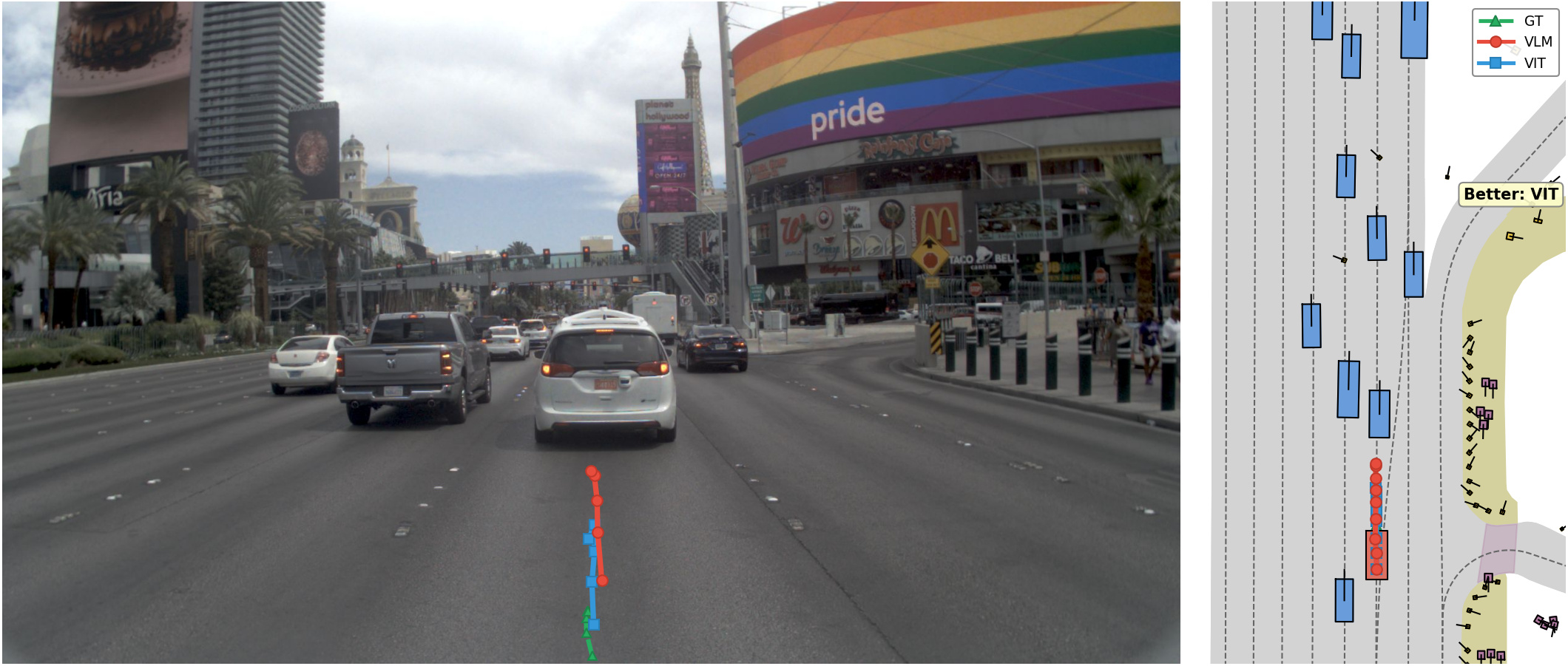}\\[-1mm]
    \includegraphics[width=\linewidth]{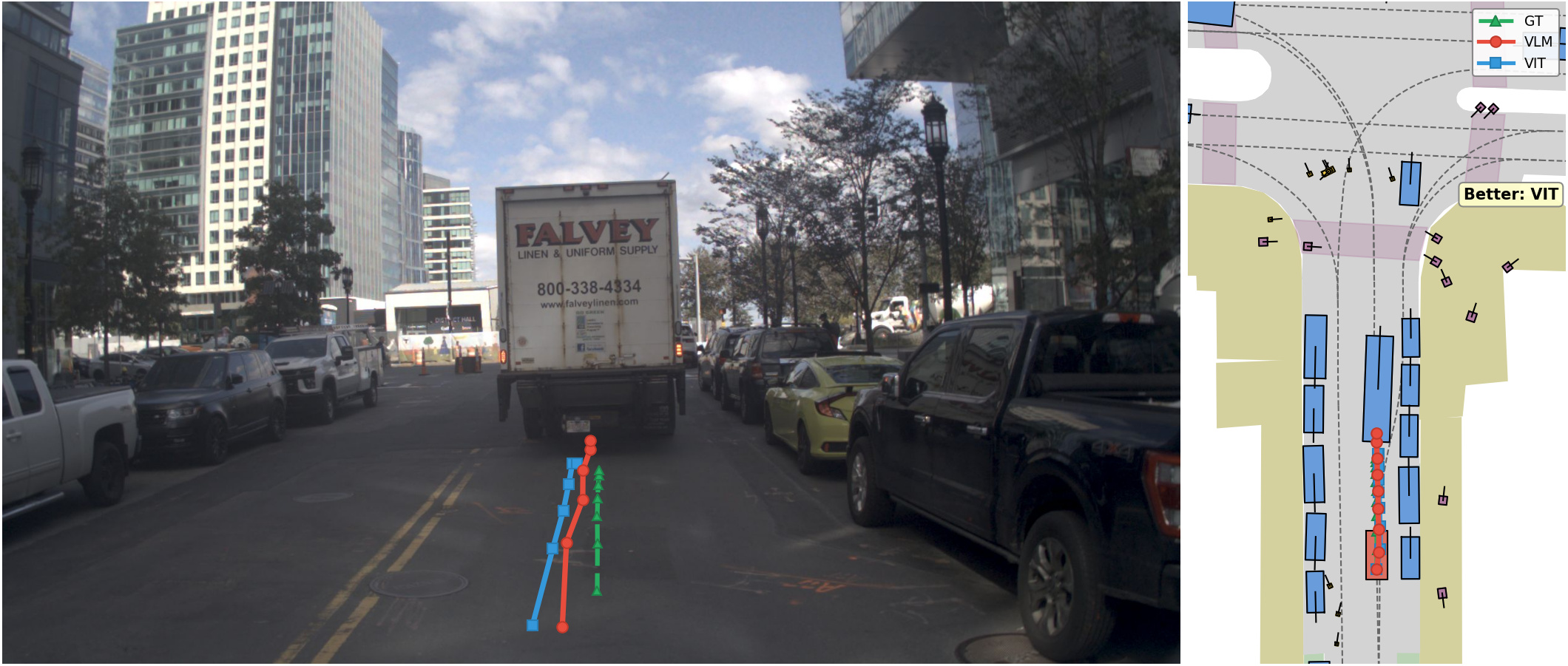}\\[-1mm]
    \includegraphics[width=\linewidth]{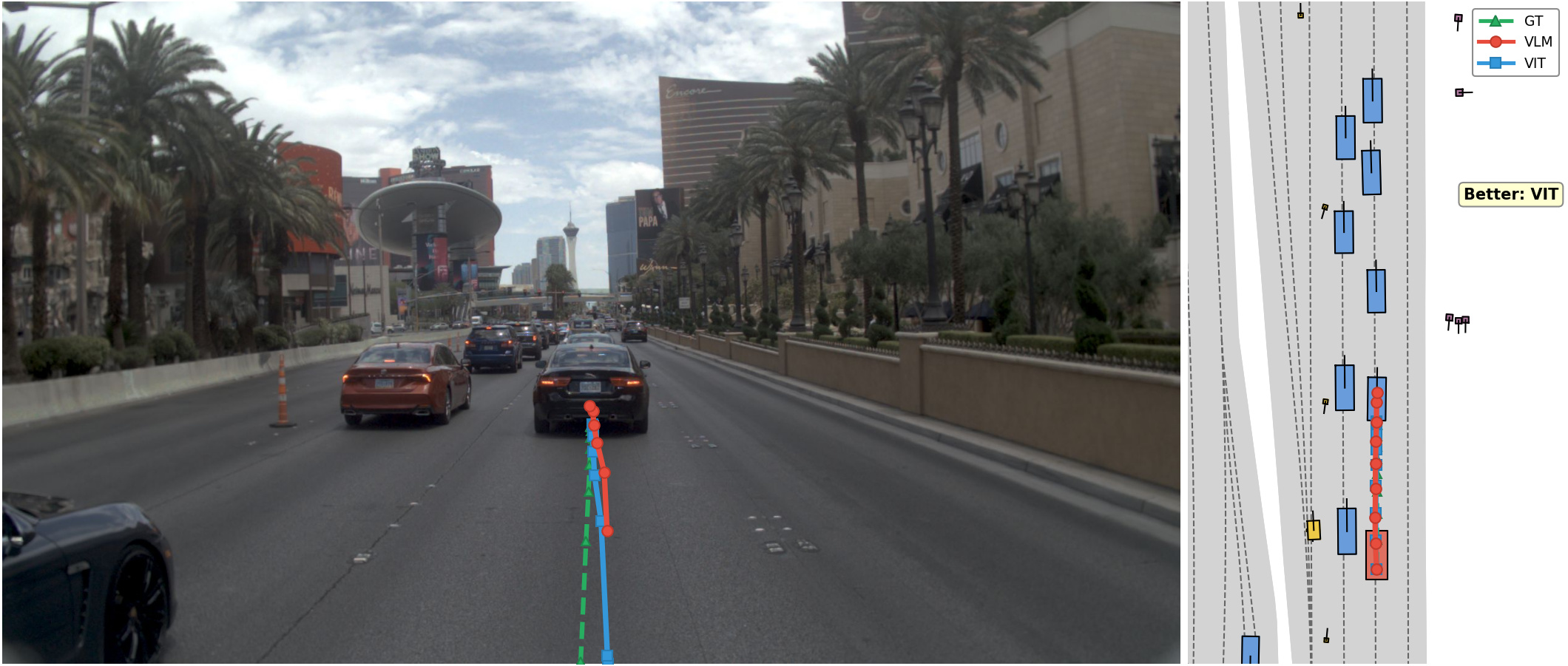}
  \end{minipage}
  \caption{Qualitative case gallery arranged in two columns. Each panel contains a front-camera view and a BEV visualization. Red denotes the VLM trajectory, blue denotes the ViT trajectory, and green denotes the human (expert) trajectory used as imitation-learning supervision. The left column (Cases 1--5) primarily highlights longitudinal / speed-profile differences, while the right column (Cases 6--10) highlights lateral / path and lane-level preference differences.}
  \label{fig:rq2_case_gallery}
\end{figure}


\end{document}